\documentclass[10pt,journal,compsoc]{IEEEtran}
\usepackage{algorithm}
\usepackage{algorithmic}
\usepackage{csquotes} 
\usepackage{color}
\usepackage{paralist}
\usepackage{latexsym}
\usepackage{indentfirst}
\usepackage{subfigure}
\usepackage{float}
\usepackage{cite}
\usepackage{CJKutf8}
\usepackage{multirow}
\usepackage{makecell}
\usepackage{mathrsfs}
\usepackage[colorlinks, linkcolor=red,  anchorcolor=blue, citecolor=blue]{hyperref}
\usepackage{textcomp,booktabs}
\usepackage{amssymb}
\usepackage{pifont}
\usepackage{epsfig}
\usepackage{colortbl}
\usepackage{amsmath} 
\usepackage{siunitx}
\usepackage{longtable}
\definecolor{mygray}{gray}{.9}
\definecolor{mypink}{rgb}{.99,.91,.95}
\definecolor{mycyan}{cmyk}{.3,0,0,0}
\hypersetup{%
        anchorcolor=blue, 
        colorlinks,
        breaklinks=true,
        plainpages=false,%
        citecolor=blue,
        linkcolor=red,
        urlcolor=magenta,
        bookmarksopen=true,%
        bookmarksnumbered=false,%
        bookmarksdepth=5%
}

\usepackage{booktabs} 
\usepackage{makecell} 
\usepackage{lipsum} 

\usepackage{colortbl}
\definecolor{mygray}{gray}{.9}
\definecolor{mypink}{rgb}{.99,.91,.95}
\definecolor{mycyan}{cmyk}{.3,0,0,0}

\usepackage{color}
\usepackage{xcolor}
\definecolor{citecolor}{HTML}{0071bc} 
\definecolor{SeaGreen4}{RGB}{0,205,102} 
\definecolor{SlateBlue}{RGB}{106,90,205} 
\definecolor{DarkRed}{RGB}{178,34,34} 
\usepackage{ragged2e}
\usepackage{graphicx}
\usepackage{soul}
\usepackage{textcomp}
\usepackage{array}
\usepackage{microtype}
\usepackage[english]{babel}

\begin{document}
\begin{CJK}{UTF8}{gbsn}



\title{Pre-training, Reasoning, Benchmarking: X-ray Report Generation on CheXpert Plus Dataset}

\author{ Xiao Wang, \textit{Member, IEEE}, Yuxiang Zhang, Dan Xu, Yuehang Li, Shiao Wang, \\ 
    Bo Jiang*, Yaowei Wang, \textit{Member, IEEE}, Yonghong Tian, \textit{Fellow, IEEE}, Jin Tang*  

\IEEEcompsocitemizethanks{
\IEEEcompsocthanksitem Xiao Wang, Yuxiang Zhang, Dan Xu, Yuehang Li, Shiao Wang, Bo Jiang, and Jin Tang are with the School of Computer Science and Technology, Anhui University, Hefei 230601, China (email: \{xiaowang, jiangbo, tangjin\}@ahu.edu.cn, z1169647007@foxmail.com, 18856245162@163.com, e23201112@stu.ahu.edu.cn, wsa1943230570@126.com) 
\IEEEcompsocthanksitem Yaowei Wang is with Peng Cheng Laboratory, Shenzhen, China; Harbin Institute of Technology, Shenzhen, China. (email: wangyw@pcl.ac.cn)  
\IEEEcompsocthanksitem Yonghong Tian is with Peng Cheng Laboratory, Shenzhen, China; National Key Laboratory for Multimedia Information Processing, School of Computer Science, Peking University, China; School of Electronic and Computer Engineering, Shenzhen Graduate School, Peking University, China. (email: yhtian@pku.edu.cn) 
\IEEEcompsocthanksitem Corresponding author: Bo Jiang \& Jin Tang 
}}


\markboth{IEEE Transactions on Pattern Analysis and Machine Intelligence}%
{Shell \MakeLowercase{\textit{et al.}}: Bare Demo of IEEEtran.cls for Computer Society Journals}

\IEEEtitleabstractindextext{%
\begin{abstract} 
\justifying 
X-ray image-based Radiology Report Generation (RRG) constitutes a critical research direction within medical artificial intelligence, with great potential to alleviate clinicians’ diagnostic workload and shorten patient waiting periods. Despite substantial advances over recent years, the field faces evident bottlenecks stemming from insufficient standardized benchmarks and inadequate domain adaptation of generic large models. Notably, the newly released CheXpert Plus dataset is provided without accompanying baseline implementations and evaluation results, which impedes standardized training, quantitative evaluation and fair comparison among follow-up algorithms. To mitigate this limitation, we establish a comprehensive benchmark encompassing prevailing X-ray report generation models and Large Language Models on CheXpert Plus. This benchmark delivers a reliable comparative foundation for upcoming methods and enables researchers to rapidly identify state-of-the-art approaches within this domain. Beyond benchmark construction, we rethink X-ray RRG under the paradigm of large models and propose a novel framework termed MambaXray-PRB. Our framework improves report generation performance and enhances model interpretability via multi-stage large-model pre-training and multi-modal Chain-of-Thought reasoning. The pipeline consists of three successive phases: self-supervised auto-regressive modeling, X-ray–report contrastive learning, and post-training optimization for reasoning and report generation. Extensive experiments on IU X-ray, MIMIC-CXR, and CheXpert Plus datasets validate the effectiveness of MambaXray-PRB for radiology report generation. 
The source code of this paper is available on \url{https://github.com/Event-AHU/Medical_Image_Analysis} 
\end{abstract}

}

\maketitle

\IEEEdisplaynontitleabstractindextext

%
\IEEEpeerreviewmaketitle

\IEEEraisesectionheading{\section{Introduction} \label{sec:introduction}} 

\IEEEPARstart{R}{adiology} Report Generation (RRG)~\cite{seah2025RRGsurvey} is a crucial task for the implementation of artificial intelligence in the medical field, which aims to automatically analyze X-ray medical imaging, identify lesion information, sort out key imaging conclusions, and generate standardized, accurate, and normative medical radiology reports through artificial intelligence algorithms. This task can greatly reduce the diagnostic pressure on doctors, effectively shorten the time required for image reading and report writing, lower the risks of manual writing errors, misdiagnosis and missed diagnosis, and improve the efficiency and homogenization level of medical imaging diagnosis. It also helps enhance the diagnostic capacity of primary medical institutions and promote the popularization of high-quality medical resources. However, there are still many unsolved challenges in this research direction, such as insufficient extraction of complex lesion features, low accuracy of rare disease image recognition, poor logicality and clinical professionalism of generated reports, and poor adaptability to report specifications and diagnosis and treatment scenarios in different hospitals. How to further improve the performance of RRG remains a worthy research problem.

\begin{figure*}
\centering
\includegraphics[width=1\linewidth]{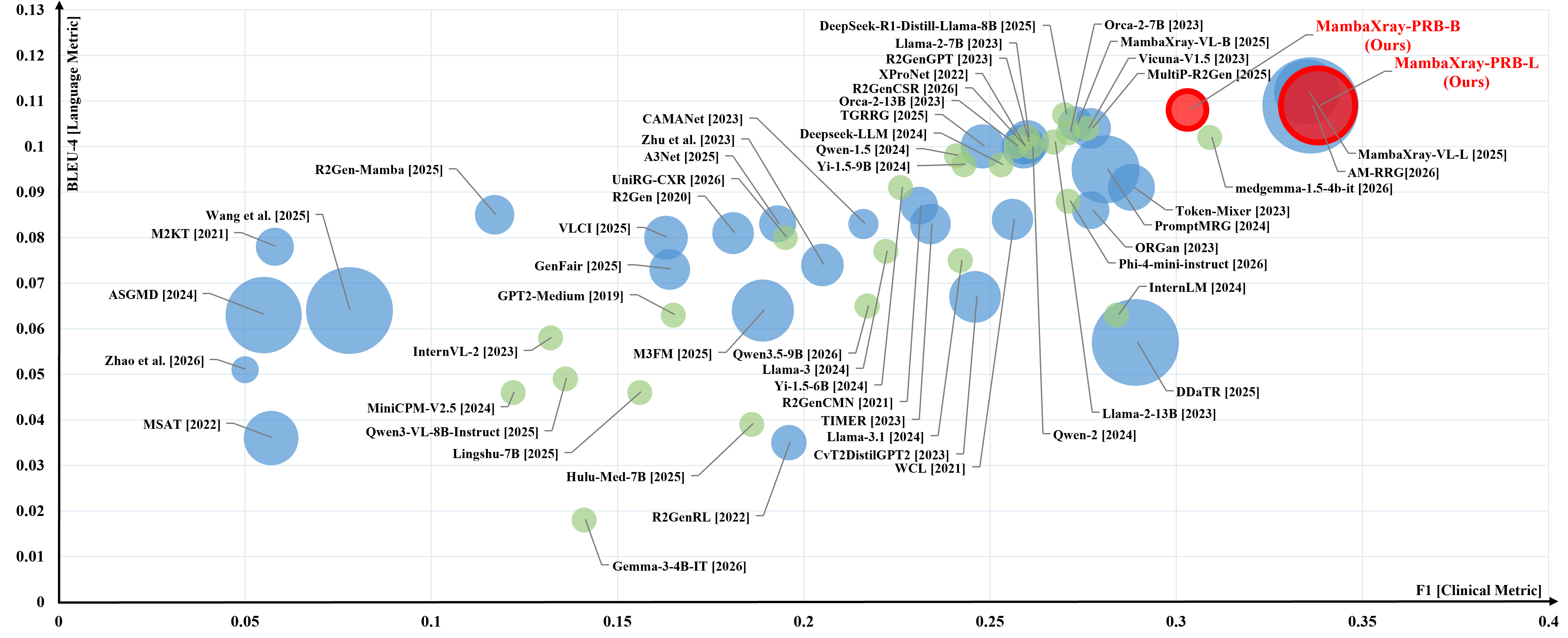}
\caption{An overview of the benchmarked LLM/VLM-based (\textcolor[RGB]{169,209,142}{green circle}) and mainstream MRG models (\textcolor[RGB]{91,155,213}{blue circle}) on the CheXpert Plus dataset in this paper. The CXPMRG-Bench integrates 30 mainstream algorithms, 16 large language models, and 9 vision-language models for X-ray medical report generation. 
} 
\label{fig:placeholder}
\end{figure*}

According to our observation, radiology report generation usually follows the encoder-decoder framework, which adopts a vision encoder (e.g., ResNet~\cite{he2016ResNet}, Transformer~\cite{NIPS2017_selfattention}) to process the given X-ray image and a language decoder (e.g., LSTM~\cite{Hochreiter1997LSTM}, GRU~\cite{Chung2014GRU}, Transformer~\cite{NIPS2017_selfattention}) for report generation. Specifically, text and image inputs are encoded using a text encoder and a visual encoder, respectively; the visual embeddings are then projected into the latent space of the text embeddings, and finally, the two are concatenated and fed into a decoder to generate the report. 
Recently, pre-trained LLMs (Large Language Models) and VLMs (Vision-Language Models) have also been introduced into the RRG community. For example, Hulu-Med~\cite{jiang2025hulumed} develops a generalist medical VLM that unifies the understanding of medical text, 2D images, 3D volumes, and videos within a single architecture, supporting a broad range of clinical tasks including medical report generation. Lingshu~\cite{lasateam2025lingshu} progressively incorporates domain-specific medical knowledge into a general-purpose VLM through multi-stage vision--language alignment and instruction tuning, with further reinforcement learning to strengthen medical reasoning capabilities.

After reviewing mainstream algorithms for X-ray medical report generation, we observe that datasets including IU X-ray~\cite{demner2016iuxray} and MIMIC-CXR~\cite{johnson2019mimicCXR} serve as the standard training and evaluation testbeds for report generation models. Nevertheless, IU X-ray merely comprises 7,470 images paired with 3,955 radiology reports, rendering its scale relatively constrained, particularly in the era of large models. The recently released CheXpert Plus dataset~\cite{chambon2024chexpertPLUS} provides a large-scale resource for X-ray report generation research. Unfortunately, the original publication does not supply reproducible baseline implementations, which hinders fair experimental comparisons among follow-up algorithms on this dataset. To address this gap, we establish CXPMRG-Bench, a comprehensive benchmark encompassing prevailing open-source X-ray report generation models, large language models (LLMs), and vision-language models (VLMs), and perform systematic evaluations on the newly released CheXpert Plus dataset, as illustrated in Table~\ref{tab:XraybaseAlgorithms} and Table~\ref{tab:xrayLLMs}. This benchmark enables researchers to quantitatively identify state-of-the-art large models and algorithms for X-ray report generation.

From another algorithmic viewpoint, we believe these RRG works are still limited by the following issues: 
1). The widely used Transformer vision backbone network brings huge computational cost $\mathcal{O}(N^2)$, which is not hardware-friendly. 
2). Many existing X-ray models~\cite{wang2024pretraininghighdefinitionxray, You2023CXR-Clip, Chen2023PTUnifier} are pre-trained on publicly available datasets with paired text annotations. Nevertheless, the limited scale of such paired data inherently constrains model performance. In contrast, pure visual X-ray data is abundant, making its effective harnessing the key to boosting model performance. 
3). Existing LLM-based radiology report generation methods~\cite{Wang2023R2GenGPT,li2024kargen,hou2025radar} primarily optimize vision--language alignment and report-generation accuracy, while providing limited explicit, stepwise interpretability grounded in image evidence.
Recent advances in Chain-of-Thought (CoT) reasoning~\cite{jing2026radiologist,jin2025chain} reveal that organically integrating visual signals into CoT text generation can yield more effective reasoning pathways, substantially improving the quality of generated reports. 
Therefore, it is natural to raise the following question: \textit{How to develop a computationally efficient radiology report generation framework capable of leveraging abundant unlabeled X-ray images and supporting visual-grounded interpretable reasoning?}

In this paper, we revisit X-ray radiology report generation in the era of large models and propose a novel \textit{MambaXray-PRB} framework, which advances report generation quality and model interpretability from the perspectives of multi-stage large-model pre-training and multi-modal Chain-of-Thought (CoT) reasoning. As shown in Fig.~\ref{fig:framework}, our framework can be separated into three stages, including \textit{self-supervised auto-regressive generation} and \textit{X-ray-report contrastive learning}, and \textit{post-training for reasoning $\&$ report generation}. Specifically, we first partition X-ray images and feed them into the Mamba network to auto-regressively predict subsequent tokens conditioned on previous contextual tokens. This design substantially strengthens X-ray visual perception capability with the efficient $\mathcal{O}(N)$ Mamba architecture at low computational cost. In the second stage, we adopt contrastive learning by feeding paired X-ray images and radiology reports into the Mamba vision backbone and text encoders (Bio-ClinicalBERT~\cite{Emily2019Bio_ClinicalBERT} and Llama2~\cite{touvron2023llama2}), which effectively aligns visual and textual features within a unified pre-trained embedding space. On top of this foundation, we perform multi-modal chain-of-thought reasoning post-training and task-specific supervised fine-tuning across diverse downstream X-ray report generation datasets. By inheriting the pre-trained knowledge and thinking steps, our framework achieves superior report generation performance on three mainstream medical report generation benchmark datasets.

To sum up, the main contributions of this paper can be summarized as follows: 

1). We propose a novel pre-trained foundation model for X-ray radiology report generation, termed \textit{MambaXray-PRB}, which adopts Mamba as the vision encoder and the large language model as the decoder. Leveraging a multi-stage pre-training strategy to maximize the efficacy of large-scale data, \textit{MambaXray-PRB} incorporates a multi-modal chain-of-thought reasoning mechanism. It produces high-quality radiology reports and explicitly characterizes the X-ray visual-linguistic reasoning procedure, thereby improving interpretability.

2). We further extend and refine a comprehensive benchmark, namely \textit{CXPMRG-Bench}, for the newly released CheXpert Plus dataset~\cite{chambon2024chexpertPLUS}. Significantly expanded in scale and coverage, the upgraded benchmark integrates 30 mainstream X-ray medical report generation algorithms, 16 large language models, and 9 vision-language models. To the best of our knowledge, this work establishes the first large-scale and systematic evaluation benchmark tailored to the CheXpert Plus dataset, and provides comprehensive and authoritative baseline references and standardized comparison criteria for subsequent research in the field of X-ray report generation.

3). We further broaden our experimental scope to the IU X-ray~\cite{demner2016iuxray} and MIMIC-CXR~\cite{johnson2019mimicCXR} datasets, and introduce the GREEN metric~\cite{ostmeier2024green} alongside traditional metrics for thorough evaluation. Supported by analytical experiments and qualitative visualizations, we deeply analyze the performance and X-ray report generation capabilities of our \textit{MambaXray-PRB} model.

A preliminary version of this work appeared in the proceedings of the IEEE/CVF Conference on Computer Vision and Pattern Recognition (CVPR) 2025. Compared with the conference predecessor, CXPMRG-Bench~\cite{wang2025cxpmrg}, this paper introduces novel and insightful contributions, including a revised methodological framework, an expanded and more comprehensive benchmark, and thorough experimental analyses and elaborations. 
Overall, in our new version, we make the following extensions: 
\textbf{(I).} A newly designed multi-modal chain-of-thought reasoning mechanism that significantly improves model interpretability. 
\textbf{(II).} A larger-scale and more comprehensive evaluation benchmark for the CheXpert Plus (CXP) dataset. 
\textbf{(III).} Sufficient experimental validations, standardized evaluation metrics, and in-depth analytical discussions.

\section{Related Work} \label{sec:RelatedWorks}

\subsection{Radiology Report Generation} 
Radiology report generation has attracted increasing attention in both artificial intelligence and clinical medicine. Recent studies have explored longitudinal modeling, external knowledge integration, retrieval augmentation, and clinically aligned optimization to improve the accuracy and reliability of generated reports. Wang et al.~\cite{wang2024HERGen} propose HERGen, which employs a group causal Transformer and curriculum learning to capture temporal dependencies across patient visits. Liu et al.~\cite{liu2025mlrg} introduce MLRG, which integrates spatial information from current multi-view images with temporal information from longitudinal examinations through contrastive learning and flexibly handles missing patient-specific prior knowledge. R2GenCSR~\cite{wang2024r2gencsr} combines a Mamba-based visual encoder with context sample retrieval to enhance visual representation and discriminative learning. CXPMRG-Bench~\cite{wang2025cxpmrg} provides a comprehensive benchmark and introduces MambaXray-VL, which adopts multi-stage pre-training for effective image--text representation learning.

To incorporate additional clinical information, KACL~\cite{Sha2025KACL} integrates medical knowledge, contrastive alignment, and historical report retrieval, while RA-RRG~\cite{Park2026RARRG} retrieves clinically relevant key phrases to reduce unsupported descriptions. MediVLM~\cite{Goswami2025MediVLM} aligns anatomical region features with textual representations through cross-attention. RadVLM-GRPO~\cite{Gundersen2026RadVLMGRPO} employs clinically grounded reinforcement learning to improve both report generation and visual grounding. Mahmood and Syeda-Mahmood~\cite{mahmood2026phrase} propose a phrase-grounded automatic preference optimization method that aligns generated findings with their corresponding image regions to improve report quality and visual grounding. RIHA~\cite{Chen2026RIHA} further establishes fine-grained image--report correspondences through hierarchical alignment at the paragraph, sentence, and word levels. Overall, recent methods have gradually evolved toward temporally aware, knowledge-enhanced, retrieval-augmented, visually grounded, and clinically reliable report generation. 

Nevertheless, existing methods primarily focus on improving cross-modal alignment, visual grounding, or generation quality, while the explicit integration of disease-relevant visual evidence into the diagnostic reasoning process remains relatively underexplored. Our framework addresses this gap by incorporating localized disease-relevant image evidence into multimodal reasoning supervision, explicitly connecting visual evidence, diagnostic reasoning, and report generation within a unified learning framework.

\subsection{Chain-of-Thought Reasoning}
Chain-of-Thought (CoT) reasoning has become increasingly prominent in NLP for enabling models to solve complex tasks through step-by-step, interpretable reasoning.
In early work, Zhang et al.~\cite{zhang2024chain} propose an optimized Chain-of-Preference method that fine-tunes language models by leveraging the inherent preference information during the tree search process.
Gradually, Chain-of-Thought reasoning has been widely applied to various multimodal tasks.
CCoT~\cite{mitra2024compositional}, a novel zero-shot thought chain method, uses scene graph representation to extract the combinatorial knowledge of LMM, to improve the performance of LMM in the field of vision and language. 
CoT-VLA~\cite{zhao2025cot} integrates explicit CoT reasoning into the Visual Language Action model by predicting future image frames as visual targets and generating a sequence of actions to achieve them.
Dong et al.~\cite{dong2025insight} propose Insight-V, a data generation method and training pipeline for long-chain reasoning in complex vision-language tasks, aiming to enhance the reasoning capabilities of multimodal large language models.
Wu et al.~\cite{wu2026grounded} introduce Grounded Chain-of-Thought, which guides multimodal large language models to progressively identify and ground relevant visual evidence during the reasoning process.
Tian et al.~\cite{tian2026reagen} propose ReaGEN, an adaptive framework that generates sample-specific structured reasoning chains to improve multimodal reasoning accuracy and efficiency.
Different from generic multimodal CoT methods, we construct radiology-specific reasoning supervision and associate intermediate diagnostic reasoning with disease-relevant image patches to support evidence-grounded report generation.

\subsection{Large Foundation Models} 
Large foundation models have demonstrated strong transfer, generation, and reasoning capabilities across natural language processing and various downstream tasks~\cite{brown2020language, chowdhery2023palm, touvron2023llama, raffel2020exploring, bai2023qwen, guo2025deepseek}. These advances have been largely driven by Transformer architectures~\cite{NIPS2017_selfattention}, large-scale training data, increased computational resources, and improved post-training strategies. GPT-3~\cite{brown2020language}, for example, shows that model scaling can substantially enhance few-shot learning without task-specific fine-tuning. More recent LLMs place greater emphasis on reasoning and efficiency. DeepSeek-R1~\cite{guo2025deepseek} employs reinforcement learning to elicit advanced reasoning behaviors, while Qwen3~\cite{yang2025qwen3} unifies thinking and non-thinking modes for adaptive inference. MiMo-V2-Flash~\cite{xiao2026mimov2flash} further combines a sparse Mixture-of-Experts architecture with efficient long-context modeling and post-training distillation.

Medical and multimodal foundation models have similarly evolved toward unified understanding across heterogeneous inputs. MedVersa~\cite{zhou2024MedVersa} uses an LLM-based orchestrator to coordinate diverse medical imaging and language tasks within a unified framework. Hulu-Med~\cite{jiang2025hulumed} integrates medical text, 2D images, 3D volumes, and videos through a shared vision encoder and LLM decoder. Qwen3.5-Omni~\cite{qwen2026qwen35omni} unifies understanding and generation across text, images, audio, and video, whereas MedGemma 1.5~\cite{sellergren2026medgemma15} extends medical foundation modeling to volumetric imaging, pathology slides, anatomical localization, and longitudinal chest X-rays. Building upon these advances, we progressively adapt the visual encoder and LLM through radiographic representation learning, cross-modal alignment, and evidence-grounded multimodal reasoning for reliable report generation.

\subsection{State Space Model} 
Transformers~\cite{NIPS2017_selfattention} have become a dominant architecture for sequence modeling, but the quadratic complexity of self-attention limits their efficiency on long sequences and high-resolution inputs. Mamba~\cite{gu2023mamba} addresses this limitation through input-dependent selective state space models (SSMs) and hardware-aware parallel scanning, enabling linear-complexity sequence modeling. The success of Mamba has motivated its extension to computer vision. Vim~\cite{zhu2024visionMamba} employs positional embeddings and bidirectional SSM blocks for visual representation learning, while VMamba~\cite{liu2024vmamba} introduces 2D selective scanning to capture global visual context with linear complexity. ARM~\cite{ren2024arm} combines Mamba-based token mixing with MLPs and auto-regressive visual pre-training. Mamba-2~\cite{Dao2024mamba2} further introduces State Space Duality, reformulating SSM computation as hardware-efficient matrix multiplication.

Recent developments continue to improve the expressiveness and scalability of SSMs. Mamba-3~\cite{lahoti2026mamba3} introduces a more expressive discretized recurrence, complex-valued state updates, and a multi-input multi-output formulation to enhance state tracking and decoding efficiency. C\textsuperscript{2}SSM~\cite{wu2026c2ssm} replaces pixel-wise scanning with cluster-centric state-space modeling, substantially reducing the cost of global dependency modeling in ultra-high-resolution images. For medical vision, GeoSemba~\cite{sun2026geosemba} reconstructs state-space information flow to jointly model geometric--semantic interactions and spatial--channel dependencies for efficient medical image segmentation. Inspired by the successful application of Mamba to diverse vision tasks~\cite{wang2024mambaevt, huang2024mambaFETrack, wang2024MambaPAR}, we employ Mamba for auto-regressive X-ray representation learning and subsequently align the learned visual features with clinical semantics for report generation. Building on the aligned representations, our framework further couples the Mamba-based visual encoder with an LLM to perform evidence-grounded clinical reasoning and radiology report generation.

\section{Methodology} \label{sec:Approach}

\subsection{Overview} 
\begin{figure*}
    \centering
    \includegraphics[width=1\linewidth]{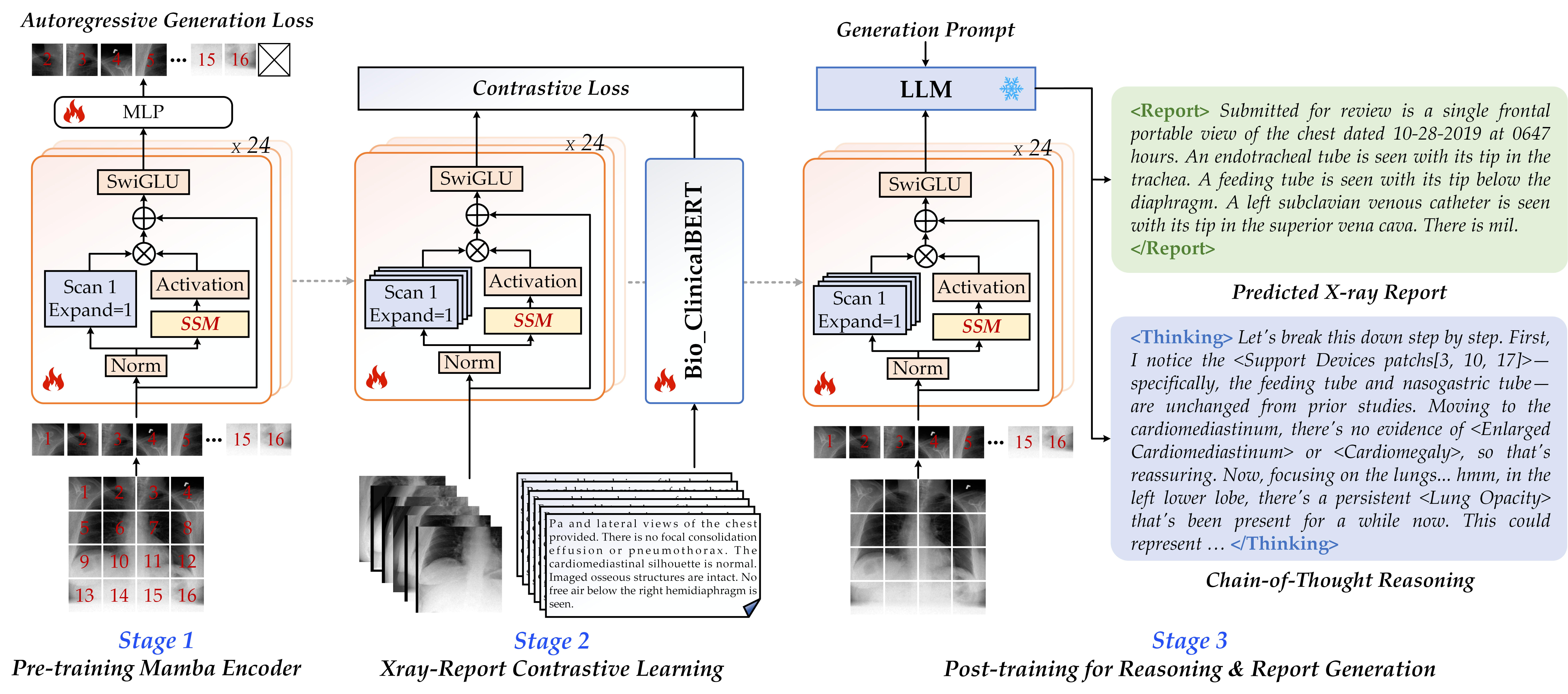}
    \caption{An overview of our proposed MambaXray-PRB pre-training and reasoning framework. It can be separated into three stages, including \textit{self-supervised auto-regressive generation} and \textit{X-ray-report contrastive learning}, and \textit{post-training for reasoning $\&$ report generation}. Specifically, the first stage aims to obtain a better visual backbone network (this paper chooses the low-complexity Mamba model) using larger-scale X-ray medical images. In the second stage, we adopt contrastive learning by feeding paired X-ray images and radiology reports into the Mamba vision backbone and text encoders, which effectively aligns visual and textual features within a unified pre-trained embedding space. The third phase is used to perform multi-modal chain-of-thought reasoning post-training and task-specific supervised fine-tuning across diverse downstream X-ray report generation datasets. Note that the layers or modules with \textit{fire/snow} symbols denote the parameters that are tuned/frozen in the training phase. }
    \label{fig:framework}
\end{figure*}

As illustrated in Fig.~\ref{fig:framework}, we formulate chest X-ray report generation as a progressive learning process that evolves from visual structure modeling to cross-modal semantic alignment and, ultimately, evidence-grounded clinical reasoning. Our framework contains three stages. First, a Mamba-based image encoder is initialized through self-supervised auto-regressive learning, in which an X-ray image is represented as an ordered patch sequence and future patches are predicted from preceding visual context. This stage equips the encoder with both local anatomical representations and long-range spatial dependencies using large-scale unpaired X-ray images. Second, paired images and reports are introduced to establish correspondence between radiographic patterns and clinical semantics. The image encoder and a clinical text encoder extract modality-specific representations, which are aligned in a shared embedding space through contrastive learning.

Compared to the conference version of the paper, we made changes to the training in the third stage. We construct explicit multimodal reasoning supervision for each training pair. An LLM converts the reference report into a clinically coherent chain of thought, while a disease recognition model localizes image regions associated with the reported abnormalities. The indices of these disease-relevant patches are inserted into the reasoning sequence, thereby linking diagnostic statements to their supporting visual evidence. In the final stage, the aligned visual encoder is coupled with an LLM and post-trained to generate the evidence-aware reasoning sequence and medical report in a unified auto-regressive process. Consequently, the model not only learns to describe imaging findings but also provides a logical reasoning process.

\subsection{Pre-Training Stage}

\noindent $\bullet$ \textbf{Stage I. Auto-regressive Generation for Mamba Vision Encoder Pre-training.~}  
To fully exploit the large-scale publicly available chest X-ray images and learn more effective visual representations before supervised training for medical report generation, we first pre-train the visual encoder. Inspired by auto-regressive generative models such as ChatGPT~\cite{openai2023chatGPT}, GPT-4~\cite{openai2024gpt4}, and ARM~\cite{ren2024arm}, we adopt an auto-regressive pre-training strategy that is well aligned with the Mamba backbone architecture to pre-train the visual module. Specifically, given an input chest X-ray image $\mathcal{I} \in \mathbb{R}^{192 \times 192 \times 3}$, we partition it into a sequence of non-overlapping image patches $\{\mathcal{P}_i\}_{i=1}^{N}$, where each patch has a size of $\mathcal{P}_i \in \mathbb{R}^{16 \times 16 \times 3}$. These image patches are projected by a convolutional embedding layer with a kernel size of $16 \times 16$ into high-dimensional visual tokens $\mathcal{T}_i \in \mathbb{R}^{1024}, i \in \{1,2,\ldots,N\}$. The embedded visual tokens are then fed into the Vim~\cite{zhu2024visionMamba} backbone for feature extraction. As illustrated in Fig.~\ref{fig:framework}, the input visual tokens are first normalized and then processed by two parallel branches, namely, the SSM branch and the scan branch. The output features are combined with the original inputs through residual connections. To enhance the non-linear representation capability, the features produced by each Mamba block are further transformed using Swish-Gated Linear Units (SwiGLU)~\cite{Shazeer2020SwiGLU}. Finally, an MLP layer is appended to predict the target visual tokens from the output features:
\begin{align}
    p(\mathcal{T}) = \prod_{i=1}^{N} p(\mathcal{T}_i | \mathcal{T}_1, ..., \mathcal{T}_{i-1}, \Theta). 
\end{align}

The visual encoder is optimized with an auto-regressive generation loss during this stage: 
\begin{align}
    \label{auto-regressiveLoss} 
    \mathcal{L}_{AR} = \sum_{i=1}^{N-1} |Vim([\mathcal{T}_1, ..., \mathcal{T}_i]) - \mathcal{T}_{i+1}|^2. 
\end{align}

The resulting Vim-based visual encoder weights are subsequently used to initialize the task-specific model in the next stage.

\noindent $\bullet$ \textbf{Stage II. Contrastive Learning in Vision-Language.~} 
In the second stage, we reuse the Mamba-based visual backbone obtained from Stage I and perform contrastive learning on paired X-ray images and medical reports. Following the contrastive learning strategy adopted in CLIP~\cite{Alec2021CLIP}, this stage aims to further improve the alignment between the visual and textual modalities. Specifically, we randomly sample a mini-batch and encode the X-ray images using the Vim backbone, while the corresponding medical reports are processed by a language model, such as Bio-ClinicalBERT~\cite{Emily2019Bio_ClinicalBERT} or Llama2~\cite{touvron2023llama2}. The cosine similarities between matched and unmatched X-ray–report pairs are then calculated as follows:
\begin{align}
    \label{eq:ctl_loss}
    \mathcal{L}_{CTL}
    =
    \operatorname{Similarity}\bigl(
        Vim(I_i),
        LM(R_j)
    \bigr),
\end{align}
where $i$ and $j$ denote the indices of the X-ray images and their report annotations, respectively.

\begin{figure*}
\centering
\includegraphics[width=1\linewidth]{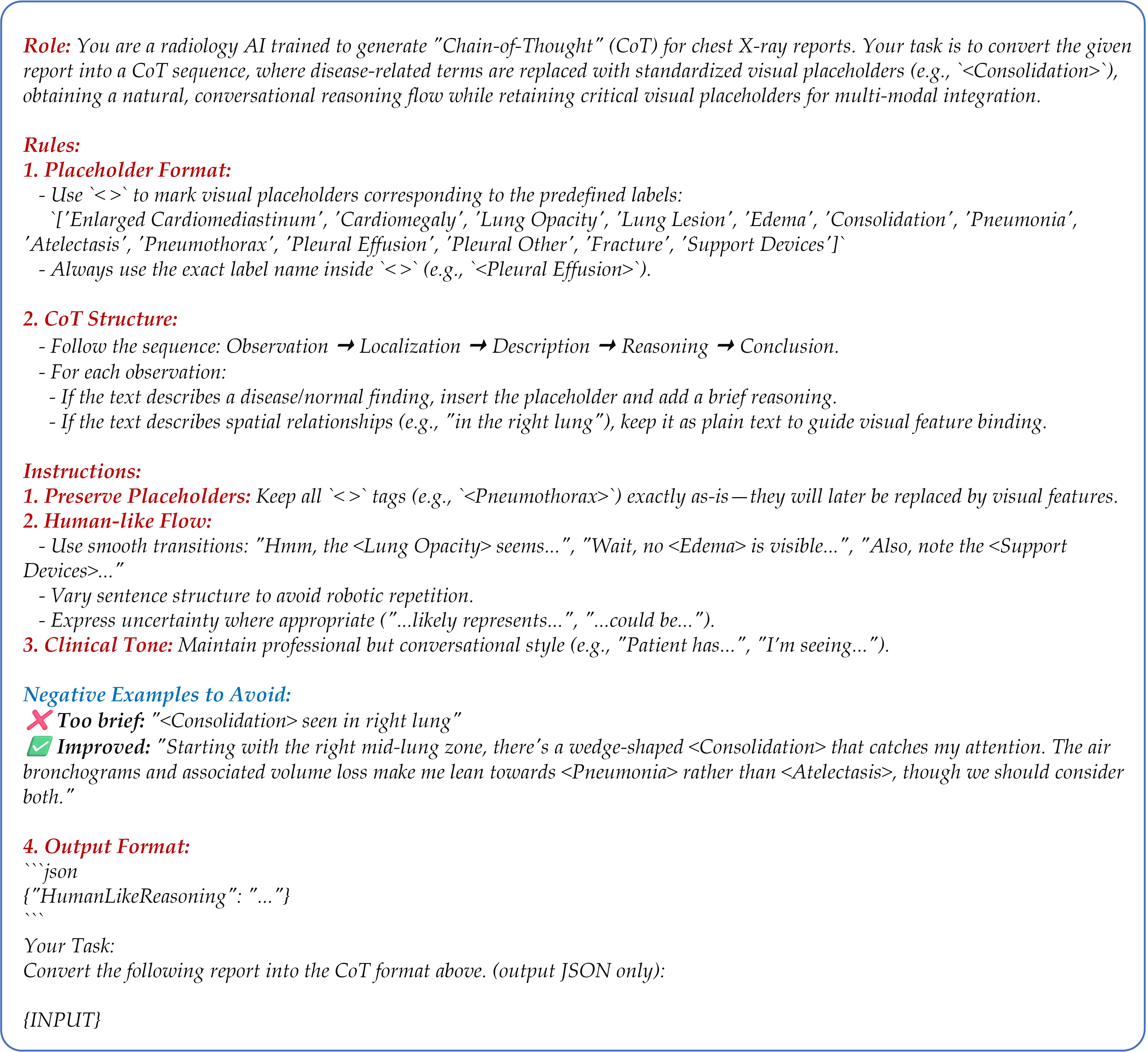}
\caption{Instruction template for constructing Chain-of-Thought (CoT) reasoning traces from chest X-ray radiology reports with standardized visual placeholders for multimodal integration.}
\label{fig:PromptCoT}
\end{figure*}

\subsection{Multimodal Reasoning Generation}

To address the scarcity of reasoning-oriented data in the medical domain, we construct a large-scale multi-modal Chain-of-Thought (CoT) dataset through an automated generation pipeline. Given the high cost and labor-intensive nature of manual annotation, our approach leverages existing radiology reports from three publicly available datasets: IU X-Ray~\cite{demner2016iuxray}, MIMIC-CXR~\cite{johnson2019mimicCXR}, and CheXpert
Plus~\cite{chambon2024chexpertPLUS}. Below, we detail the key steps of our methodology.


\noindent $\bullet$ \textbf{CoT Generation via LLM Prompting.~} Given an input chest X-ray image $\mathcal{I}$ and its corresponding radiology report $\mathcal{R}$, we employ a structured prompt template to guide a large language model (LLM) in generating clinically grounded reasoning chains. This process formalizes the CoT generation as a mapping function $\mathcal{G}: \mathcal{R} \rightarrow \mathcal{C}$, where $\mathcal{C}$ represents the structured CoT sequence. The template enforces standardized disease placeholders (e.g., $\langle \text{Consolidation} \rangle$) while preserving natural language flow through a five-stage reasoning sequence: Observation $\rightarrow$ Localization $\rightarrow$ Description $\rightarrow$ Reasoning $\rightarrow$
Conclusion. The report $\mathcal{R}$ is fed into LLM with our CoT template to generate $\mathcal{C}_{\text{raw}}$.

\noindent $\bullet$ \textbf{Visual Information Extraction.~} To establish visual-linguistic grounding, we first partition $\mathcal{I} \in \mathbb{R}^{H \times W \times 1}$ into $N$ non-overlapping patches $\{\mathcal{P}_k\}_{k=1}^{N}$ of size $16 \times 16$ pixels, where $N = \lfloor H/16 \rfloor \times \lfloor W/16 \rfloor$. Patches are indexed in row-major order with global index $k = i \cdot \lfloor W/16 \rfloor + j$ for spatial position $(i,j)$.
For each pathology $l \in \mathcal{L}$ (the predefined set of 13 thoracic diseases), we generate class activation maps $\mathcal{M}_l$ using a disease encoder $\mathcal{V}$~\cite{wang2025cxpmrg}. The activation score $\alpha_k^{(l)}$ for patch $P_k$ is computed through spatial averaging:
\begin{equation}
\alpha_k^{(l)} = \underset{(x,y)\in P_k}{\mathrm{mean}} \mathcal{M}_l(x,y).
\end{equation}

We then select relevant patches using a threshold $\tau$ determined for each disease $l$. The selected patch set $\mathcal{S}_l$ is defined as:
\begin{equation}
\mathcal{S}_l = \left\{ k \in \{1,\dots,N\} \mid \alpha_k^{(l)} \geq \tau \right\}.
\end{equation}

\begin{figure*}
\centering
\includegraphics[width=1\linewidth]{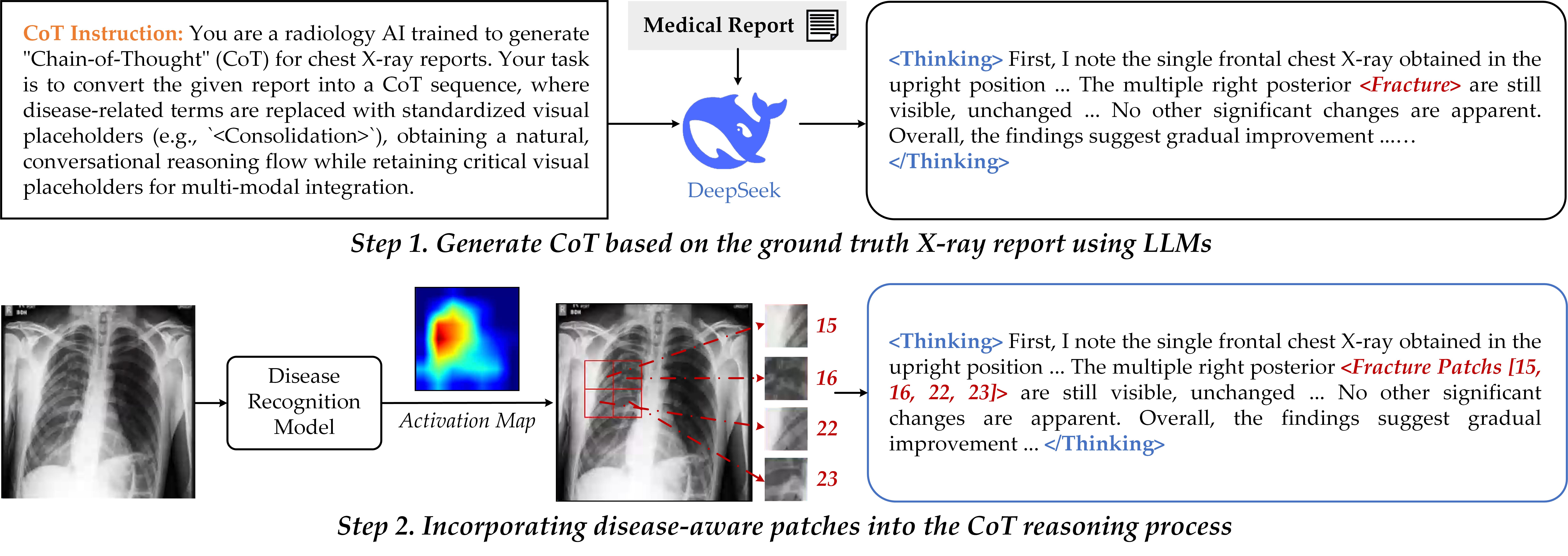}
\caption{The Generation Process of Visual-aware Multi-modal Chain-of-Thought Reasoning Data. In the first stage, we adopt ground-truth reports as prompts to guide the large language model (we employ the DeepSeek model in this work) to generate high-quality text-only Chain-of-Thought reasoning. In the second stage, we leverage a disease classification model to extract highly activated regions from X-ray images. These visual cues are fused with the text-based Chain-of-Thought obtained in the first stage to construct multi-modal Chain-of-Thought data. }
\label{fig:CoT_data}
\end{figure*}

The final CoT $\mathcal{C}$ integrates visual localization through:
\begin{equation}
\mathcal{C} = \Psi \left( \mathcal{C}_{\text{raw}}, \{\mathcal{S}_l\}_{l \in \mathcal{L}} \right),
\end{equation}
where $\Psi$ replaces each occurrence of $\langle l \rangle$ with $\langle l[\mathcal{S}_l] \rangle$ using deterministic string substitution. For
example, a phrase "right upper lobe $\langle \text{Consolidation} \rangle$" becomes "right upper lobe $\langle \text{Consolidation patches}[42,43,58,59] \rangle$", explicitly linking the pathological finding to patches 42, 43, 58, and 59 in the image grid. This tight visual-linguistic coupling enables downstream models to associate textual reasoning steps with spatially grounded evidence.

\subsection{Post-Training Stage}

To optimize our framework, we formulate the training objective using a multi-task learning paradigm. Given an input X-ray image $\mathcal{I}$ and its corresponding CoT sequence $\mathcal{C}$, we first extract visual tokens using a vision encoder $\mathcal{V}$:

\begin{equation}
\mathbf{V} = \mathcal{V}(\mathcal{I}) \in \mathbb{R}^{d_v \times N_v},
\end{equation}
where $d_v$ denotes the visual feature dimension and $N_v$ the number of visual tokens. 
These visual tokens are projected into the language model's embedding space via a linear layer to attain $\mathbf{H}_v$.

The model processes the combined input sequence $\mathbf{X} = [\mathbf{H}_v, \mathbf{H}_t]$ using an auto-regressive language model. Here, $\mathbf{H}_t$ represents the prompt embedding. We define two specialized loss functions:

\noindent $\bullet$ \textbf{Thinking Branch Loss.~} 
Computed for tokens generated after $\langle \text{thinking} \rangle$  tags:
\begin{equation}
\mathcal{L}_{th} = -\sum_{t=1}^{T_{\text{th}}} \log p(c_t | \mathbf{X}, c_{<t}; \theta),
\end{equation}
where $c_t$ is the $t$-th token in the ground-truth CoT sequence $\mathcal{C} = \{c_1, c_2, \dots, c_{T_{\text{th}}}\}$, and $\theta$ denotes model parameters.

\noindent $\bullet$ \textbf{Answer Branch Loss.~} 
Computed for tokens after $\langle \text{answer} \rangle$:
\begin{equation}
\mathcal{L}_{\text{ans}} = -
\sum_{t=1}^{T_{\text{ans}}} \log p(r_t | \mathbf{X},  r_{<t}; \theta),
\end{equation}
where $r_t$ is the $t$-th token in the ground-truth report $\mathcal{R} = \{r_1, r_2, \dots, r_{T_{\text{ans}}}\}$.

The final training objective combines both losses with balancing coefficients:
\begin{equation}
\mathcal{L}_{\text{total}} = \alpha \mathcal{L}_{\text{th}} + \beta \mathcal{L}_{\text{ans}},
\end{equation}
where $\alpha$ and $\beta$ control the relative importance of CoT reasoning versus report generation. Unless otherwise specified, both are set to 1.0 by default.

During inference, the model supports three distinct generation modes through flexible prompt engineering:

\begin{align}
\hat{\mathcal{C}} & = \text{LLM}(\mathbf{X} | \langle \text{thinking} \rangle; \theta), \\
\hat{\mathcal{R}} & = \text{LLM}(\mathbf{X} | \langle \text{answer} \rangle; \theta).
\end{align}

This unified architecture maintains parameter consistency while accommodating diverse clinical needs: radiologists can extract justifications for complex cases through pure CoT generation, obtain efficient summaries for routine exams via direct reporting, or request complete diagnostic rationales with evidence chains when necessary, all within the same model instance.

\section{CheXpert Plus Benchmark} \label{sec:CXPBench}

\subsection{Mainstream X-ray Report Generation Models}
As reported in Table~\ref{tab:XraybaseAlgorithms}, we compare our approach with a broad range of representative X-ray medical report generation (MRG) models. These methods cover different visual architectures, including CNN-, Transformer-, and Mamba-based encoders, together with recurrent, Transformer-based, or large language model decoders. The selected baselines represent mainstream developments in radiology report generation and provide a comprehensive comparison with our MambaXray-PRB models on the CheXpert Plus dataset.

For a fair evaluation, we use the complete reference reports when computing all metrics. During the reproduction of existing methods, we observed that some implementations truncate the ground-truth reports before evaluation, which may lead to inconsistent comparisons across models. We therefore remove such truncation and evaluate all reproduced methods against the full ground truth under a unified protocol. This setting ensures that the reported results reflect the generation quality of different models under the same evaluation conditions.

\begin{table*}
\centering
\caption{Experimental results on the CheXpert Plus dataset using mainstream medical report generation methods. 
\textbf{B4}, \textbf{R-L}, \textbf{M}, and \textbf{C} denote BLEU-4, ROUGE-L, METEOR, and CIDEr, respectively, while \textbf{P}, \textbf{R}, and \textbf{F1} denote Precision, Recall, and F1 score. \textit{min} denotes minutes. Param denotes the number of trainable parameters during training. The best and second-best performance values are highlighted in bold and underlined, respectively.}
\label{tab:XraybaseAlgorithms}
\small
\resizebox{\textwidth}{!}{
\begin{tabular}{c|l|l|l|l|l|l|c|c|c}
\hline \toprule [0.5 pt]
\textbf{Index} & \textbf{Algorithm} & \textbf{Publish} & \textbf{Encoder} & \textbf{Decoder}
& \textbf{\hspace{1ex}B4,\hspace{1em}R-L,\hspace{1em}M,\hspace{1em}C}
& \textbf{\hspace{2ex}P,\hspace{1em}\hspace{1ex}R,\hspace{1em}\hspace{1ex}F1}
& \textbf{\makecell{Time \\ (\textit{min})}}
& \textbf{\makecell{Param \\ (\textit{M})}}
& \textbf{Code} \\
\hline
\#01 & R2Gen~\cite{Chen2020R2Gen} & EMNLP20 & Transformer & Transformer
& 0.081, 0.246, 0.113, 0.077 & 0.318, 0.200, 0.181 & 110.05 & 83.50 & \hyperlink{https://github.com/zhjohnchan/R2Gen}{URL} \\
\hline
\#02 & M2KT~\cite{Yang2023M2KT} & MIA21 & CNN & Transformer
& 0.078, 0.247, 0.101, 0.077 & 0.044, 0.142, 0.058 & 22.50 & 69.07 & \hyperlink{https://github.com/LX-doctorAI1/M2KT}{URL} \\
\hline
\#03 & R2GenCMN~\cite{Chen2021R2GenCMN} & ACL21 & Transformer & Transformer
& 0.087, 0.256, 0.127, 0.102 & 0.329, 0.241, 0.231 & 66.08 & 67.70 & \hyperlink{https://github.com/zhjohnchan/R2GenCMN}{URL} \\
\hline
\#04 & WCL~\cite{yan2021WCL} & EMNLP21 & Transformer & Transformer
& 0.084, 0.253, 0.126, 0.103 & 0.335, 0.259, 0.256 & 24.08 & 81.29 & \hyperlink{https://github.com/zzxslp/WCL}{URL} \\
\hline
\#05 & R2GenRL~\cite{qin2022R2GenRL} & ACL22 & Transformer & Transformer
& 0.035, 0.186, 0.101, 0.012 & 0.193, 0.229, 0.196 & 44.33 & 59.87 & \hyperlink{https://github.com/synlp/R2GenRL}{URL} \\
\hline

\#06 & XProNet~\cite{wang2022XProNet} & ECCV22 & Transformer & Transformer
& 0.100, 0.265, 0.146, 0.121 & 0.314, 0.247, 0.259 & 6.30 & 62.35 & \hyperlink{https://github.com/Markin-Wang/XProNet}{URL} \\
\hline

\#07 & MSAT~\cite{wang2022MSAT} & MICCAI22 & ViT-B/16 & Transformer
& 0.036, 0.156, 0.066, 0.018 & 0.044, 0.142, 0.057 & 5.72 & 141.10 & \hyperlink{https://github.com/wang-zhanyu/MSAT}{URL} \\
\hline

\#08 & ORGan~\cite{hou2023ORGan} & ACL23 & CNN & Transformer
& 0.086, 0.261, 0.135, 0.107 & 0.288, 0.287, 0.277 & 46.66 & 67.50 & \hyperlink{https://github.com/wjhou/ORGan}{URL} \\
\hline

\#09 & TIMER~\cite{Wu2023TIMER} & CHIL23 & Transformer & Transformer
& 0.083, 0.254, 0.121, 0.104 & 0.345, 0.238, 0.234 & 26.50 & 79.28 & \hyperlink{https://github.com/woqingdoua/TIMER}{URL} \\
\hline

\#10 & CvT2DistilGPT2~\cite{Nicolson2023CvT2DistilGPT2} & AIM23 & Transformer & GPT2
& 0.067, 0.238, 0.118, 0.101 & 0.285, 0.252, 0.246 & 13.93 & 128.00 & \hyperlink{https://github.com/aehrc/cvt2distilgpt2}{URL} \\
\hline

\#11 & Zhu et al.~\cite{Zhu2023LongitudinalChestX-Rays} & MICCAI23 & Transformer & Transformer
& 0.074, 0.235, 0.128, 0.078 & 0.217, 0.308, 0.205 & 10.03 & 85.95 & \hyperlink{https://github.com/CelestialShine/Longitudinal-Chest-X-Ray}{URL} \\
\hline

\#12 & CAMANet~\cite{wang2024CAMANet} & IEEE JBH23 & Swin-Former & Transformer
& 0.083, 0.249, 0.118, 0.090 & 0.328, 0.224, 0.216 & 23.08 & 43.22 & \hyperlink{https://github.com/Markin-Wang/CAMANet}{URL} \\
\hline

\#13 & Token-Mixer~\cite{Yang2024Token-Mixer} & IEEE TMI23 & ResNet-50 & Transformer
& 0.091, 0.261, 0.135, 0.098 & 0.309, 0.270, 0.288 & 17.54 & 104.34 & \hyperlink{https://github.com/yangyan22/Token-Mixer}{URL} \\
\hline

\#14 & R2GenGPT~\cite{Wang2023R2GenGPT} & Meta-Rad.23 & Swin-Transformer & Llama2
& 0.101, 0.266, 0.145, 0.123 & 0.315, 0.244, 0.260 & 77.80 & 90.90 & \hyperlink{https://github.com/wang-zhanyu/R2GenGPT}{URL} \\
\hline

\#15 & ASGMD~\cite{XUE2024ASGMD} & ESWA24
& \makecell[l]{ResNet-101\\Transformer} & Transformer
& 0.063, 0.220, 0.094, 0.044 & 0.146, 0.108, 0.055 & 87.37 & 277.41 & \hyperlink{https://github.com/shangchengLu/ASGMDN}{URL} \\
\hline

\#16 & PromptMRG~\cite{Jin2024PromptMRG} & AAAI24 & ResNet-101 & Bert
& 0.095, 0.222, 0.121, 0.044 & 0.258, 0.265, 0.281 & 108.45 & 219.92 & \hyperlink{https://github.com/jhb86253817/PromptMRG}{URL} \\
\hline

\#17 & VLCI~\cite{chen2024VLCI} & IEEE TIP25 & Transformer & Transformer
& 0.080, 0.247, 0.114, 0.072 & 0.341, 0.175, 0.163 & 123.71 & 91.46 & \hyperlink{https://github.com/WissingChen/VLCI}{URL} \\
\hline

\#18 & Wang et al.~\cite{wang2024pretraininghighdefinitionxray} &  Visual Intelligence25 & ViT & Llama2
& 0.064, 0.220, 0.110, 0.059 & 0.175, 0.099, 0.078 & 10.82 & 358.80 & \hyperlink{https://github.com/aehrc/cxrmate}{URL} \\
\hline

\#19 & MultiP-R2Gen~\cite{Chen2025MultiPR2Gen} & TMI25 & Swin-Transformer & Llama2
& 0.104, 0.169, \textbf{0.269}, \underline{0.140} & 0.318, 0.246, 0.277 & 36.18 & 78.90 & \hyperlink{https://github.com/tylin/coco-caption}{URL} \\
\hline

\#20 & A3Net~\cite{Nguyen2025A3Net} & ICIP25 & Transformer & Transformer
& 0.083, 0.248, 0.124, 0.088 & 0.287, 0.206, 0.193 & 89.72 & 65.16 & \hyperlink{https://github.com/Vinh-AI/A3Net}{URL} \\
\hline

\#21 & GenFair~\cite{Chen2025GenFair} & NCS25 & ResNet & Transformer
& 0.073, 0.237, 0.107, 0.067 & 0.325, 0.181, 0.164 & 76.68 & 79.28 & \hyperlink{https://github.com/iriscxy/GenFair}{URL} \\
\hline

\#22 & R2Gen-Mamba~\cite{Sun2025R2GenMamba} & IEEE 22nd ISBI & Mamba & Transformer
& 0.085, 0.248, 0.119, 0.066 & 0.218, 0.163, 0.117 & 2.83 & 74.55 & \hyperlink{https://github.com/CHARAN422176/R2Gen-Mamba-master}{URL} \\
\hline

\#23 & TGRRG~\cite{Guangli2025TGRRG} & MedIA25
& \makecell[l]{CNN\\Transformer} & Transformer
& 0.100, 0.264, 0.147, 0.099 & 0.333, 0.228, 0.248 & 34.77 & 92.44 & \hyperlink{https://github.com/chentaohuang/Topic-Guided-Radiology-Report-Generation}{URL} \\
\hline

\#24 & M3FM~\cite{Liu2025M3FM} & npj D.M.25 & Bert & Bert
& 0.064, 0.221, 0.128, 0.039 & 0.250, 0.247, 0.189 & 2020.40 & 183.22 & \hyperlink{https://github.com/ai-in-health/M3FM}{URL} \\
\hline

\#25 & DDaTR~\cite{Song2025DDaTR} & IEEE TMI25 & \makecell[l]{Swin-Transformer\\Bert} & Bert
& 0.057, 0.205, 0.118, 0.031 & 0.314, 0.317, 0.289 & 229.32 & 360.95 & \hyperlink{https://github.com/xmed-lab/DDaTR}{URL} \\
\hline

\#26 & MambaXray-VL-B~\cite{wang2025cxpmrg} & CVPR25 & MambaXray-VL & Llama2
& 0.105, 0.267, 0.149, 0.117 & 0.333, 0.264, 0.273 & 50.66 & 57.31 & \hyperlink{https://github.com/Event-AHU/Medical_Image_Analysis}{URL} \\
\hline

\#27 & MambaXray-VL-L~\cite{wang2025cxpmrg} & CVPR25 & MambaXray-VL & Llama2
& \textbf{0.112}, \underline{0.276}, 0.157, 0.139 & 0.377, \textbf{0.319}, 0.335 & 55.18 & 202.32 & \hyperlink{https://github.com/Event-AHU/Medical_Image_Analysis}{URL} \\
\hline

\#28 & Zhao et al.~\cite{Zhao2026MPAF} & IEEE ICASSP26
& \makecell[l]{ResNet-101\\Transformer} & Transformer & 0.051, 0.186, 0.082, 0.019 & 0.106, 0.143, 0.050 & 4.18 & 34.93 & \hyperlink{https://github.com/liuhh-code/mycode}{URL} \\
\hline

\#29 & R2GenCSR~\cite{wang2024r2gencsr} & IEEE JBHI26 & VMamba & Llama2
& 0.100, 0.265, 0.146, 0.121 & 0.315, 0.247, 0.259 & 31.20 & 91.70 & \hyperlink{https://github.com/Event-AHU/Medical_Image_Analysis/tree/main/R2GenCSR}{URL} \\
\hline

\#30 & AM-MRG~\cite{wang2026ammrg} & IEEE TMI26 & Swin-Transformer & Llama2
& \underline{0.109}, \textbf{0.282}, \underline{0.173}, \textbf{0.221} & \textbf{0.396}, \underline{0.318}, \underline{0.336} & 192.85 & 441.70 & \hyperlink{https://github.com/Event-AHU/Medical_Image_Analysis}{URL} \\
\hline


\#31 & MambaXray-PRB-B & Ours & MambaXray-VL & Llama2
& 0.108, 0.271, 0.153, 0.130 & 0.359, 0.285, 0.303 & 265.33 & 87.40 & \hyperlink{https://github.com/Event-AHU/Medical_Image_Analysis}{URL} \\
\hline

\#32 & MambaXray-PRB-L & Ours & MambaXray-VL & Llama2
& \underline{0.109}, 0.274, 0.155, 0.134 & \underline{0.387}, 0.317, \textbf{0.338} & 406.63 & 300.00 & \hyperlink{https://github.com/Event-AHU/Medical_Image_Analysis}{URL} \\
\hline \toprule [0.5 pt] 
\end{tabular}
}
\end{table*}

\begin{table*}[t]
\centering
\caption{\textsc{Green Metrics of MambaXray-PRB in MIMIC-CXR and CheXpert Plus. M.F. is short for Matched Findings.}} 
\label{tab:green_metrics}
\setlength{\tabcolsep}{5pt}
\renewcommand{\arraystretch}{1.05}
\resizebox{\textwidth}{!}{%
\begin{tabular}{c|lcccccccccc}
\toprule
\textbf{Dataset}
& \textbf{Method}
& \multicolumn{6}{c}{\textbf{Clinically Significant Errors} $\downarrow$}
& $\displaystyle\sum_{j=(a)}^{(f)} \textbf{Error}_{j}\ \downarrow$
& \textbf{M.F.} $\uparrow$
& \textbf{GREEN} $\uparrow$ \\
\cmidrule(lr){3-8}
&
& (a) & (b) & (c) & (d) & (e) & (f)
& & & \\
\midrule
\multirow{6}{*}{\textbf{MIMIC-CXR}}
& R2Gen-2020~\cite{Chen2020R2Gen} & 1.310 & 3.089 & \textbf{0.103} & \textbf{0.201} & 0.082 & 0.142 & 4.926 & 1.803 & 0.283 \\

& R2GenCMN-2021~\cite{Chen2021R2GenCMN} & 1.383 & 2.963 & 0.127 & 0.228 & \textbf{0.081} & 0.161 & 4.942 & 1.935 & 0.297 \\

& CvT2DistilGPT2-2023~\cite{Nicolson2023CvT2DistilGPT2} & 1.150 & 2.881 & 0.146 & 0.234 & 0.103 & 0.153 & 4.666 & 1.967 & 0.313 \\

& SEI-2024~\cite{kang2024sei} & 1.391 & 2.610 & 0.154 & 0.273 & 0.108 & 0.132 & 4.668 & 2.101 & 0.326 \\

& AM-MRG-2026~\cite{wang2026ammrg} & 1.636 & 2.222 & 0.158 & 0.289 & 0.140 & \textbf{0.088} & 4.533 & 2.104 & \textbf{0.330} \\

& R2GenCSR-2026~\cite{wang2024r2gencsr} & 1.577 & 2.256 & 0.156 & 0.311 & 0.144 & 0.098 & 4.541 & 2.099 & 0.329 \\

& CX-Mind-2026~\cite{LI2026CXMind} & - & - & - & - & - & - & - & - & 0.267 \\

& DM-RRG-2026~\cite{JU2026DMRRG} & - & - & - & - & - & - & - & - & 0.302 \\

\cmidrule(lr){2-11}

& MambaXray-PRB-B (Ours) & \textbf{0.057} & \textbf{0.097} & 0.840 & 0.767 & 0.834 & 0.969 & \textbf{3.564} & \textbf{2.147} & 0.304 \\

\midrule

\multirow{6}{*}{\textbf{CheXpert Plus}}
& R2Gen-2020~\cite{Chen2020R2Gen} & 1.235 & 3.638 & \textbf{0.150} & \textbf{0.122} & \textbf{0.021} & 0.103 & 5.269 & 0.938 & 0.168 \\

& R2GenCMN-2021~\cite{Chen2021R2GenCMN} & 1.512 & 3.336 & 0.218 & 0.165 & 0.043 & 0.126 & 5.399 & 1.179 & 0.197 \\

& CvT2DistilGPT2-2023~\cite{Nicolson2023CvT2DistilGPT2} & 1.815 & 3.289 & 0.252 & 0.182 & 0.091 & 0.132 & 5.760 & 1.166 & 0.185 \\

& R2GenGPT-2023~\cite{Wang2023R2GenGPT} & 1.928 & 2.361 & 0.280 & 0.262 & 0.108 & \textbf{0.059} & 4.998 & 1.521 & 0.244 \\

& AM-MRG-2026~\cite{wang2026ammrg} & 1.935 & 2.084 & 0.273 & 0.266 & 0.145 & 0.063 & 4.768 & 1.754 & \textbf{0.281} \\

& R2GenCSR-2026~\cite{wang2024r2gencsr} & 2.015 & 2.311 & 0.258 & 0.251 & 0.121 & 0.069 & 5.024 & 1.535 & 0.246 \\

& CX-Mind-2026~\cite{LI2026CXMind} & - & - & - & - & - & - & - & - & 0.179 \\

\cmidrule(lr){2-11}

& MambaXray-PRB-B (Ours) & \textbf{0.083} & \textbf{0.048} & 0.739 & 0.743 & 0.916 & 0.899 & \textbf{3.428} & \textbf{1.805} & 0.272 \\

\bottomrule
\end{tabular}%
}
\end{table*}

\begin{table*}[]
\caption{Comparison of our model's performance on the IU X-ray and MIMIC-CXR datasets. The symbol $\dagger$ indicates that we follow the R2Gen annotation using \textit{Findings} and evaluate with our method, as their report modifies the ground truth to an \textit{Impression} concatenated with \textit{Findings}. The best result is highlighted in bold, and the second-best result is underlined.}
\label{tab:results_iu_mimic}
\resizebox{\linewidth}{!}{
\begin{tabular}{c|l|c|ccccccc}
\hline \toprule [0.5 pt] 
\textbf{Dataset} & \textbf{Methods} & \textbf{Publication} 
& \textbf{BLEU-1} & \textbf{BLEU-2} & \textbf{BLEU-3} 
& \textbf{BLEU-4} & \textbf{ROUGE-L} & \textbf{METEOR} 
& \textbf{CIDEr} \\ 
\hline

\multirow{18}{*}{\rotatebox[origin=c]{90}{\textbf{IU X-Ray}}}
 & R2Gen~\cite{Chen2020R2Gen} 
 & EMNLP 2020 
 & 0.470 & 0.304 & 0.219 & 0.165 & 0.371 & 0.187 & -- \\

 & R2GenCMN~\cite{Chen2021R2GenCMN} 
 & ACL-IJCNLP 2021 
 & 0.475 & 0.309 & 0.222 & 0.170 & 0.375 & 0.191 & -- \\

 & PPKED~\cite{liu2021PPKED} 
 & CVPR 2021 
 & 0.483 & 0.315 & 0.224 & 0.168 & 0.376 & 0.187 & 0.351 \\

 & AlignTrans~\cite{you2021aligntransformer} 
 & MICCAI 2021 
 & 0.484 & 0.313 & 0.225 & 0.173 & 0.379 & 0.204 & -- \\

 & CMCL~\cite{liu2021CMCL} 
 & ACL 2021 
 & 0.473 & 0.305 & 0.217 & 0.162 & 0.378 & 0.186 & -- \\

 & Clinical-BERT~\cite{clinicalBert} 
 & AAAI 2022 
 & 0.495 & 0.330 & 0.231 & 0.170 & 0.376 & 0.209 & 0.432 \\

 & METransformer~\cite{Wang2023METransformer} 
 & CVPR 2023 
 & 0.483 & 0.322 & 0.228 & 0.172 & 0.380 & 0.192 & 0.435 \\

 & DCL~\cite{Li2023DCL} 
 & CVPR 2023 
 & -- & -- & -- & 0.163 & 0.383 & 0.193 & \textbf{0.586} \\

 & R2GenGPT\textsuperscript{$\dagger$}~\cite{Wang2023R2GenGPT} 
 & Meta Radiology 2023 
 & 0.465 & 0.299 & 0.214 & 0.161 & 0.376 & 0.219 & 0.542 \\

 & PromptMRG~\cite{Jin2024PromptMRG} 
 & AAAI 2024 
 & 0.401 & -- & -- & 0.098 & 0.160 & \textbf{0.281} & -- \\ 

 & BootstrappingLLM~\cite{liu2024bootstrapping} 
 & AAAI 2024 
 & 0.499 & 0.323 & 0.238 & 0.184 & 0.390 & 0.208 & -- \\ 

 & MultiP-R2Gen~\cite{Chen2025MultiPR2Gen} 
 & TMI 2025 
 & 0.515 & 0.332 & 0.238 & 0.178 & 0.386 & 0.216 & 0.477 \\

 & DDaTR~\cite{Song2025DDaTR} 
 & TMI 2025 
 & 0.421 & -- & -- & 0.103 & 0.307 & 0.160 & -- \\

 & TGRRG~\cite{Guangli2025TGRRG} 
 & MedIA 2025 
 & \textbf{0.521} & \textbf{0.363} & \underline{0.265} 
 & 0.198 & \textbf{0.409} & 0.222 & 0.482 \\

 & MambaXray-VL-Base~\cite{wang2025cxpmrg} 
 & CVPR 2025 
 & 0.479 & 0.322 & 0.236 & 0.179 & 0.388 & 0.215 & 0.508 \\

 & MambaXray-VL-Large~\cite{wang2025cxpmrg} 
 & CVPR 2025 
 & 0.491 & 0.330 & 0.241 & 0.185 & 0.371 & 0.216 & 0.524 \\

 & Zhao et al.~\cite{Zhao2026MPAF} & IEEE ICASSP26
 & 0.485 & 0.317 & 0.230 & 0.178 & 0.380 & - & - \\
 
 & R2GenCSR~\cite{wang2024r2gencsr} & IEEE JBHI26
 & 0.514 & 0.351 & 0.262 
 & \textbf{0.206} & \underline{0.401} & 0.215 & \underline{0.579} \\

 & DTrace~\cite{Ye2026DTrace} & IEEE TMM26
 & \underline{0.516} & \underline{0.353} & \textbf{0.278} 
 & \underline{0.204} & 0.386 & \underline{0.233} & 0.469 \\
 
 \cline{2-10} 

 & MambaXray-PRB-B 
 & Ours 
 & 0.492 & 0.332 & 0.243 & 0.187 & 0.366 & 0.124 & 0.494 \\

 & MambaXray-PRB-L 
 & Ours 
 & 0.489 & 0.336 & 0.250
 & 0.195 & 0.377 & 0.125 & 0.527 \\

\hline \toprule [0.5 pt] 

\multirow{18}{*}{\rotatebox[origin=c]{90}{\textbf{MIMIC-CXR}}}
 & R2Gen~\cite{Chen2020R2Gen} 
 & EMNLP 2020 
 & 0.353 & 0.218 & 0.145 & 0.103 & 0.277 & 0.142 & -- \\

 & R2GenCMN~\cite{Chen2021R2GenCMN} 
 & ACL-IJCNLP 2021 
 & 0.353 & 0.218 & 0.148 & 0.106 & 0.278 & 0.142 & -- \\

 & PPKED~\cite{liu2021PPKED} 
 & CVPR 2021 
 & 0.360 & 0.224 & 0.149 & 0.106 & 0.284 & 0.149 & 0.237 \\

 & AlignTrans~\cite{you2021aligntransformer} 
 & MICCAI 2021 
 & 0.378 & 0.235 & 0.156 & 0.112 & 0.283 & 0.158 & -- \\

 & CMCL~\cite{liu2021CMCL} 
 & ACL 2021 
 & 0.344 & 0.217 & 0.140 & 0.097 & 0.281 & 0.133 & -- \\

 & Clinical-BERT~\cite{clinicalBert} 
 & AAAI 2022 
 & 0.383 & 0.230 & 0.151 & 0.106 & 0.275 & 0.144 & 0.151 \\

 & METransformer~\cite{Wang2023METransformer} 
 & CVPR 2023 
 & 0.386 & 0.250 & 0.169 & 0.124 & 0.291
 & 0.152 & \textbf{0.362} \\

 & DCL~\cite{Li2023DCL} 
 & CVPR 2023 
 & -- & -- & -- & 0.109 & 0.284 & 0.150 & 0.281 \\

 & R2GenGPT\textsuperscript{$\dagger$}~\cite{Wang2023R2GenGPT} 
 & Meta Radiology 2023 
 & 0.408 & 0.256 & 0.174 & 0.125 & 0.285 & 0.167 & 0.244 \\

 & PromptMRG~\cite{Jin2024PromptMRG} 
 & AAAI 2024 
 & 0.398 & -- & -- & 0.112 & 0.268 & 0.157 & -- \\ 

 & BootstrappingLLM~\cite{liu2024bootstrapping} 
 & AAAI 2024 
 & 0.402 & 0.262 & 0.180 & 0.128 & 0.291
 & \textbf{0.175} & -- \\ 

 & MultiP-R2Gen~\cite{Chen2025MultiPR2Gen} 
 & TMI 2025 
 & \underline{0.425} & \textbf{0.279} & \textbf{0.194} 
 & \textbf{0.140} & \underline{0.307} & 0.167 & 0.284 \\

 & DDaTR~\cite{Song2025DDaTR} 
 & TMI 2025 
 & 0.401 & -- & -- & 0.113 & 0.272 & 0.162 & -- \\

 & TGRRG~\cite{Guangli2025TGRRG} 
 & MedIA 2025 
 & \textbf{0.447} & \underline{0.276} & 0.181 & 0.124 
 & 0.291 & \underline{0.174} & 0.172 \\

 & MambaXray-VL-Base~\cite{wang2025cxpmrg} 
 & CVPR 2025 
 & 0.420 & 0.264 & 0.180 & 0.129 & 0.283 & 0.162 & 0.206 \\ 

 & MambaXray-VL-Large~\cite{wang2025cxpmrg} 
 & CVPR 2025 
 & 0.422 & 0.268 & 0.184 & 0.133
 & 0.289 & 0.167 & 0.241 \\ 

 & Zhao et al.~\cite{Zhao2026MPAF} & IEEE ICASSP26
 & 0.398 & 0.247 & 0.167 & 0.117 & 0.296 & - & - \\
 
 & R2GenCSR~\cite{wang2024r2gencsr} & IEEE JBHI26
 & 0.420 & 0.268 & \underline{0.186} & \underline{0.136} & 0.291 & 0.167 & 0.267 \\

 & DTrace~\cite{Ye2026DTrace} & IEEE TMM26
 & 0.392 & 0.260 & 0.171 & 0.129 & \textbf{0.309} & 0.162 & \underline{0.311} \\

 \cline{2-10} 

 & MambaXray-PRB-B 
 & Ours 
 & 0.404 & 0.252 & 0.171 & 0.123 & 0.284 & 0.163 & 0.231 \\

 & MambaXray-PRB-L 
 & Ours 
 & 0.412 & 0.251 & 0.166 & 0.115 & 0.276 & 0.169 & 0.229 \\

\hline \toprule [0.5 pt] 
\end{tabular}
}
\end{table*}

\begin{table}[htbp]
\centering
\footnotesize
\caption{Comparison of Clinical Efficacy (CE) metrics on MIMIC-CXR. The best result is highlighted in bold, and the second-best result is underlined. \textbf{Avg.} denotes the averaging method used; \textbf{Macro} denotes macro-averaging; \textbf{Unclear} indicates that the calculation method is unspecified.}
\label{tab:ce_comparison}
\resizebox{\linewidth}{!}{ 
\begin{tabular}{l l l c c c}
\toprule
\textbf{Method} & \textbf{Publish} & \textbf{Avg.} & \textbf{P} & \textbf{R} & \textbf{F1} \\
\midrule
R2Gen~\cite{Chen2020R2Gen}             & EMNLP 2020 & Macro & 0.333 & 0.273 & 0.276 \\
METTransformer~\cite{Wang2023METransformer}    & CVPR 2023  & Unclear & 0.364 & 0.309 & 0.311 \\
KiUT~\cite{huang2023kiut}              & CVPR 2023  & Unclear & 0.371 & 0.318 & 0.321 \\
MedRAT~\cite{hirsch2025medrat}            & ECCV 2024  & Unclear & 0.285 & 0.265 & 0.227 \\
CXR-IRGen~\cite{Shentu2024CXR-IRGen}         & WACV 2024  & Unclear & --    & --    & 0.293 \\
HERGen~\cite{wang2024HERGen}            & ECCV 2024  & Unclear & 0.415 & 0.301 & 0.317 \\
TGRRG~\cite{Guangli2025TGRRG}             & MedIA 2025 & Macro & 0.339 & \textbf{0.357} & \textbf{0.359} \\
MambaXray-VL-L~\cite{wang2025cxpmrg}    & CVPR 2025  & Macro & 0.371 & 0.321 & 0.340 \\
IMT-CXR~\cite{Xu2026IMTCXR}           & Med 2026 & Unclear & \underline{0.428} & 0.315 & 0.331 \\
RadVLM-GRPO~\cite{Gundersen2026RadVLMGRPO}            & MIDL 2026 & Macro & - & - & 0.336 \\
FedTAR~\cite{Zhu2026FedTAR}            & CVPR 2026 & Macro & \textbf{0.527} & 0.288 & 0.332 \\
MambaXray-PRB-L           &Ours         & Macro & 0.375 & \underline{0.334} & \underline{0.348} \\
\bottomrule
\end{tabular}} 
\end{table}

\subsection{LLMs and VLMs}  

Compared with the conference version, we substantially expand the model benchmark from 16 to 25 open-source foundation models, including 16 LLMs and 9 VLMs, as summarized in Table~\ref{tab:xrayLLMs}. For the LLM-based comparison, we retain the models evaluated in the conference version and additionally introduce DeepSeek-R1-Distill-Llama-8B~\cite{guo2025deepseek} and Phi-4-mini-instruct~\cite{microsoft2025phi4}. Following the same protocol, each LLM replaces the original Llama2 decoder in R2GenGPT~\cite{Wang2023R2GenGPT}, while the Swin Transformer visual encoder remains unchanged to ensure a controlled comparison.

More importantly, the journal version extends the VLM benchmark from InternVL-2~\cite{chen2023internvl} and MiniCPM-V2.5~\cite{yao2024minicpmv} to nine models by further including Qwen3-VL-8B-Instruct~\cite{yang2025qwen3}, Gemma-3-4B-IT~\cite{gemmateam2025gemma3}, UniRG-CXR~\cite{liu2026UniRGCXR}, MedGemma-1.5-4B-IT~\cite{sellergren2026medgemma15}, Hulu-Med-7B~\cite{jiang2025hulumed}, Lingshu-7B~\cite{lasateam2025lingshu}, and Qwen3.5-9B~\cite{qwen3-5}. Among them, MedGemma-1.5-4B-IT, Hulu-Med-7B, Lingshu-7B, and UniRG-CXR are medical-domain models. This expanded evaluation enables a broader comparison among language-only models, general-purpose VLMs, and medically specialized VLMs for chest X-ray report generation.


\begin{table*}
\centering
\caption{Experimental results of medical report generation on the CheXpert Plus dataset with different LLMs and VLMs under the R2Gen-GPT framework. The symbol $\dagger$ denotes a VLM. Param indicates the number of parameters of the corresponding LLM/VLM. The best result is highlighted in bold, and the second-best result is underlined.}
\label{tab:xrayLLMs}
\small
\resizebox{\textwidth}{!}{
\begin{tabular}{c|l|l|cccc|ccc|c|c|c}
\hline
\toprule[0.5pt]
\textbf{Index} & \textbf{LLM/VLM} & \textbf{Year}
& \textbf{B4} & \textbf{R-L} & \textbf{M} & \textbf{C}
& \textbf{P} & \textbf{R} & \textbf{F1}
& \textbf{Time (\textit{min})} & \textbf{Param} & \textbf{Code} \\
\hline

\#01 & Vicuna-V1.5~\cite{zheng2023vicuna} & 2023
& \underline{0.104} & \underline{0.272} & \underline{0.160} & \underline{0.202}
& 0.334 & 0.258 & 0.276
& 72.00 & 6.7B
& \hyperlink{https://huggingface.co/lmsys/vicuna-7b-v1.5}{URL} \\
\hline

\#02 & Qwen-1.5~\cite{bai2023qwen} & 2024
& 0.098 & 0.262 & 0.139 & 0.139
& 0.303 & 0.233 & 0.241
& 154.25 & 7.7B
& \hyperlink{https://huggingface.co/Qwen/Qwen1.5-7B-Chat}{URL} \\
\hline

\#03 & Qwen-2~\cite{bai2023qwen} & 2024
& 0.100 & 0.270 & 0.142 & 0.159
& 0.313 & 0.269 & 0.261
& 103.33 & 7.6B
& \hyperlink{https://huggingface.co/Qwen/Qwen2-7B-Instruct}{URL} \\
\hline

\#04 & InternLM~\cite{cai2024internlm2} & 2024
& 0.063 & 0.207 & 0.136 & 0.104
& 0.307 & \underline{0.274} & \underline{0.284}
& 294.00 & 7.3B
& \hyperlink{https://huggingface.co/internlm/internlm-chat-7b}{URL} \\
\hline

\#05 & Llama-2-7B~\cite{touvron2023llama2} & 2023
& 0.102 & 0.267 & 0.157 & 0.179
& 0.315 & 0.244 & 0.260
& 77.78 & 6.7B
& \hyperlink{https://huggingface.co/meta-llama/Llama-2-7b-chat-hf}{URL} \\
\hline

\#06 & Llama-2-13B~\cite{touvron2023llama2} & 2023
& 0.101 & 0.269 & \underline{0.160} & \textbf{0.214}
& 0.321 & 0.254 & 0.267
& 116.42 & 13.0B
& \hyperlink{https://huggingface.co/meta-llama/Llama-2-13b-chat-hf}{URL} \\
\hline

\#07 & Llama-3~\cite{dubey2024llama3} & 2024
& 0.077 & 0.220 & 0.121 & 0.134
& 0.306 & 0.232 & 0.222
& 130.00 & 8.0B
& \hyperlink{https://huggingface.co/meta-llama/Meta-Llama-3-8B-Instruct}{URL} \\
\hline

\#08 & Llama-3.1~\cite{dubey2024llama3} & 2024
& 0.075 & 0.221 & 0.121 & 0.136
& 0.295 & 0.251 & 0.242
& 110.00 & 8.0B
& \hyperlink{https://huggingface.co/meta-llama/Meta-Llama-3.1-8B-Instruct}{URL} \\
\hline

\#09 & GPT2-Medium~\cite{radford2019gpt} & 2019
& 0.063 & 0.198 & 0.104 & 0.067
& \underline{0.358} & 0.186 & 0.165
& 57.33 & 354M
& \hyperlink{https://huggingface.co/openai-community/gpt2-medium}{URL} \\
\hline

\#10 & Orca-2-7B~\cite{mitra2023orca} & 2023
& 0.103 & 0.270 & \textbf{0.161} & 0.199
& 0.330 & 0.251 & 0.271
& 177.33 & 6.7B
& \hyperlink{https://huggingface.co/microsoft/Orca-2-7b}{URL} \\
\hline

\#11 & Orca-2-13B~\cite{mitra2023orca} & 2023
& 0.100 & 0.266 & 0.159 & 0.187
& 0.317 & 0.242 & 0.257
& 108.66 & 13.0B
& \hyperlink{https://huggingface.co/microsoft/Orca-2-13b}{URL} \\
\hline

\#12 & Deepseek-LLM~\cite{deepseekai2024deepseekllm} & 2024
& 0.096 & 0.268 & 0.137 & 0.150
& 0.336 & 0.256 & 0.253
& 201.30 & 6.9B
& \hyperlink{https://huggingface.co/deepseek-ai/deepseek-llm-7b-chat}{URL} \\
\hline

\#13 & Yi-1.5-6B~\cite{ai2024yi} & 2024
& 0.091 & 0.263 & 0.131 & 0.136
& 0.322 & 0.229 & 0.226
& 43.66 & 6.1B
& \hyperlink{https://huggingface.co/01-ai/Yi-1.5-6B-Chat}{URL} \\
\hline

\#14 & Yi-1.5-9B~\cite{ai2024yi} & 2024
& 0.096 & 0.269 & 0.138 & 0.155
& 0.336 & 0.241 & 0.243
& 48.50 & 8.8B
& \hyperlink{https://huggingface.co/01-ai/Yi-1.5-9B-Chat}{URL} \\
\hline

\#15 & InternVL-2\textsuperscript{$\dagger$}~\cite{chen2023internvl} & 2023
& 0.058 & 0.188 & 0.112 & 0.085
& 0.196 & 0.127 & 0.132
& 108.50 & 8.0B
& \hyperlink{https://huggingface.co/OpenGVLab/InternVL2-8B}{URL} \\
\hline

\#16 & MiniCPM-V2.5\textsuperscript{$\dagger$}~\cite{yao2024minicpmv} & 2024
& 0.046 & 0.177 & 0.085 & 0.076
& 0.254 & 0.152 & 0.122
& 51.50 & 8.4B
& \hyperlink{https://huggingface.co/openbmb/MiniCPM-Llama3-V-2_5}{URL} \\
\hline

\#17 & DeepSeek-R1-Distill-Llama-8B~\cite{guo2025deepseek} & 2025
& \textbf{0.107} & 0.259 & 0.152 & 0.096
& 0.314 & 0.262 & 0.270
& 127.08 & 8.1B
& \hyperlink{https://huggingface.co/deepseek-ai/DeepSeek-R1-Distill-Llama-8B}{URL} \\
\hline

\#18 & Qwen3-VL-8B-Instruct\textsuperscript{$\dagger$}~\cite{yang2025qwen3} & 2025
& 0.049 & 0.212 & 0.091 & 0.071
& 0.171 & 0.135 & 0.136
& 77.87 & 8.8B
& \hyperlink{https://hf-mirror.com/Qwen/Qwen3-VL-8B-Instruct}{URL} \\
\hline

\#19 & Hulu-Med-7B\textsuperscript{$\dagger$}~\cite{jiang2025hulumed} & 2025
& 0.039 & 0.158 & 0.118 & 0.031
& 0.196 & 0.188 & 0.186
& 56.85 & 8B
& \hyperlink{https://huggingface.co/ZJU-AI4H/Hulu-Med-7B}{URL} \\
\hline

\#20 & Lingshu-7B\textsuperscript{$\dagger$}~\cite{lasateam2025lingshu} & 2025
& 0.046 & 0.218 & 0.087 & 0.068
& 0.241 & 0.206 & 0.156
& 63.20 & 8.3B
& \hyperlink{https://huggingface.co/lingshu-medical-mllm/Lingshu-7B}{URL} \\
\hline

\#21 & UniRG-CXR\textsuperscript{$\dagger$}~\cite{liu2026UniRGCXR} & 2026
& 0.080 & 0.227 & 0.130 & 0.061
& 0.234 & 0.189 & 0.195
& 91.42 & 8.8B
& \hyperlink{https://huggingface.co/microsoft/UniRG-CXR}{URL} \\
\hline

\#22 & medgemma-1.5-4b-it\textsuperscript{$\dagger$}~\cite{sellergren2026medgemma15} & 2026
& 0.102 & \textbf{0.278} & 0.147 & 0.173
& \textbf{0.387} & \textbf{0.299} & \textbf{0.309}
& 180.00 & 4.3B
& \hyperlink{https://huggingface.co/google/medgemma-1.5-4b-it}{URL} \\
\hline

\#23 & Phi-4-mini-instruct~\cite{microsoft2025phi4} & 2026
& 0.088 & 0.245 & 0.157 & 0.049
& 0.294 & 0.268 & 0.271
& 95.92 & 3.9B
& \hyperlink{https://huggingface.co/microsoft/Phi-4-mini-instruct}{URL} \\
\hline

\#24 & Qwen3.5-9B\textsuperscript{$\dagger$}~\cite{qwen3-5} & 2026
& 0.065 & 0.206 & 0.118 & 0.032
& 0.291 & 0.224 & 0.217
& 457.67 & 9.4B
& \hyperlink{https://huggingface.co/Qwen/Qwen3.5-9B}{URL} \\
\hline

\#25 & Gemma-3-4B-IT\textsuperscript{$\dagger$}~\cite{gemmateam2025gemma3} & 2026
& 0.018 & 0.168 & 0.098 & 0.011
& 0.177 & 0.175 & 0.141
& 274.25 & 12.0B
& \hyperlink{https://huggingface.co/google/gemma-3-4b-it}{URL} \\
\hline
\toprule[0.5pt]
\end{tabular}
}
\end{table*}

\section{Experiments} \label{sec:Experiments}

\subsection{Datasets and Evaluation Metrics}  

In this paper, we conduct validation on three widely used benchmark datasets, including \textbf{MIMIC-CXR}~\cite{johnson2019mimicCXR}, \textbf{IU X-ray}~\cite{demner2016iuxray}, \textbf{CheXpert Plus}~\cite{chambon2024chexpertPLUS} datasets. A brief introduction to these datasets is given below. 

\noindent $\bullet$ \textbf{MIMIC-CXR Dataset~}~\cite{johnson2019mimicCXR}~\footnote{\url{https://physionet.org/content/mimic-cxr/2.0.0/}}
 is one of the largest publicly accessible chest X-ray datasets with free-text radiology reports. It includes 377,110 radiographic images and 227,835 reports from 65,379 patients, collected between 2011 and 2016 at the Beth Israel Deaconess Medical Center in Boston. For a consistent comparison with previous methods, we adopt the dataset split used by R2GenGPT~\cite{Wang2023R2GenGPT} and exclude samples containing anomalous chain-of-thought sequences. This results in 270,686 training samples, 2,130 validation samples, and 3,858 test samples.

\noindent $\bullet$ \textbf{IU X-ray Dataset~}~\cite{demner2016iuxray}~\footnote{\url{https://iuhealth.org/find-medical-services/x-rays}}
 is a widely adopted benchmark for medical report generation. Released in 2016, it contains 7,470 radiographic images and 3,955 reports, where each report corresponds to either a frontal image or a pair of frontal and lateral images. The reports are organized into four sections: Indication, Comparison, \textit{Findings}, and \textit{Impression}. Following R2GenGPT~\cite{Wang2023R2GenGPT}, we partition the dataset into training, validation, and testing sets using a 7:2:1 ratio.

\noindent $\bullet$ \textbf{CheXpert Plus Dataset~}~\cite{chambon2024chexpertPLUS}~\footnote{\url{https://github.com/Stanford-AIMI/chexpert-plus}}
 is a large-scale radiology dataset developed to improve the scalability, robustness, and fairness of radiological deep learning systems. It provides 223,228 chest X-rays in DICOM and PNG formats, 187,711 de-identified radiology reports parsed into 11 sections, demographic information from 64,725 patients, 14 pathology labels, and RadGraph~\cite{Saahil2021RadGraph} annotations. Following R2GenCSR~\cite{wang2024r2gencsr}, we use the Findings section as the target report and divide the data into training, validation, and test subsets in a ratio of 7:1:2. We also exclude samples containing anomalous chain-of-thought sequences. The resulting subsets contain 40,441, 5,780, and 11,562 samples, respectively.

\noindent $\bullet$ \textbf{Evaluation Metrics.~}
We evaluate report quality using four widely adopted natural language generation metrics: \textbf{CIDEr}~\cite{vedantam_cider_2015}, \textbf{BLEU}~\cite{papineni_bleu_2002}, \textbf{ROUGE-L}~\cite{lin_rouge_2004}, and \textbf{METEOR}~\cite{banerjee_meteor_2005}. These metrics assess the similarity between generated and reference reports from complementary perspectives, including n-gram overlap, TF-IDF-weighted consensus, longest common subsequences, and semantic matching.

To assess the correctness of clinically relevant findings, we additionally report \textbf{Clinical Efficacy (CE)} metrics. The CheXpert labeler~\cite{irvin2019chexpert} is applied to generated and reference reports to extract major clinical observations, which are compared using \textbf{Precision}, \textbf{Recall}, and \textbf{F1-score}. Furthermore, we employ \textbf{GREEN}(Generative Radiology Report Evaluation and Error Notation)~\cite{ostmeier2024green} to provide a more fine-grained assessment of clinical consistency by identifying and characterizing clinically significant errors in generated reports. GREEN complements lexical- and label-based metrics with an expert-aligned score and interpretable error annotations.

\subsection{Implementation Details}  

\noindent $\bullet$ \textbf{Pre-training Stage.}
MambaXray-PRB was pre-trained in two phases. In the first phase, the Base and Large variants were trained for 100 epochs with batch sizes of 256 and 128, respectively. The base learning rate was set to \(1.5 \times 10^{-4}\) for a batch size of 256, and AdamW~\cite{loshchilov2019AdamW} with a weight decay of 0.05 was used together with a cosine decay schedule and a five-epoch warm-up. Input images were resized to \(192 \times 192\). Following ARM~\cite{ren2024arm}, we adopted unidirectional scanning to fit the auto-regressive generation objective and improve pre-training efficiency. In the second phase, vision--text contrastive learning was performed on 480,000 image--text pairs collected from MIMIC-CXR~\cite{johnson2019mimicCXR}, CheXpert Plus~\cite{chambon2024chexpertPLUS}, and IU X-ray~\cite{demner2016iuxray} to align visual representations with the textual feature space. The scanning block was expanded to four copies to enhance multidirectional visual modeling. This phase lasted 50 epochs with a batch size of 192. Vim~\cite{zhu2024visionMamba}, initialized with the visual weights learned in the first phase, and Bio-ClinicalBERT~\cite{Emily2019Bio_ClinicalBERT} were used as the visual and text encoders, respectively, with both encoders jointly optimized. The optimizer settings remained unchanged, while the input resolution was increased to \(224 \times 224\). During the construction of the multimodal chain-of-thought, we experimented with extracting image patch indices from images at two different resolutions, \(192 \times 192\) and \(224 \times 224\).

\noindent $\bullet$ \textbf{Fine-tuning Stage.}
We evaluated MambaXray-PRB on IU X-ray~\cite{demner2016iuxray}, MIMIC-CXR~\cite{johnson2019mimicCXR}, and CheXpert Plus~\cite{chambon2024chexpertPLUS}. For IU X-ray, training was conducted for up to 30 epochs with a batch size of 20. The visual encoder was initialized from the second-stage pre-trained Vim~\cite{zhu2024visionMamba} checkpoint, while Qwen-1.5-1.8B~\cite{bai2023qwen} was used as the language model. The maximum report length was 60 tokens, and validation was performed after every epoch. For the Base model, the reasoning sequence was limited to 120 tokens and each disease label includes a maximum of six referenced patch indices; for the Large model, these limits were increased to 180 tokens and 12 patch indices. For MIMIC-CXR and CheXpert Plus, the maximum numbers of training epochs were set to 15 and 10, respectively, with a batch size of 6. The visual encoder remained unchanged, whereas Llama2-7B~\cite{touvron2023llama2} was adopted as the language model. The maximum report length was set to 100 tokens, and validation was conducted twice per epoch. On MIMIC-CXR, the maximum reasoning lengths were 200 tokens for the Base model and 700 tokens for the Large model, with both variants allowing up to 12 patch indices for each disease label. On CheXpert Plus, both models used a maximum output reasoning length of 200 tokens and allowed up to 12 patch indices for each disease label. Except for the MIMIC-CXR Large setting, where the language model was adapted using LoRA, the language model was kept frozen throughout downstream training, and only the visual encoder and visual mapping layer were optimized following R2GenGPT~\cite{Wang2023R2GenGPT}.

\subsection{Comparison on Public Benchmark Datasets}   

\noindent $\bullet$ \textbf{Results on MIMIC Dataset.~}
As shown in Table~\ref{tab:results_iu_mimic}, MambaXray-PRB maintains competitive report-generation performance on MIMIC-CXR~\cite{johnson2019mimicCXR}. Among the compared approaches, TGRRG~\cite{Guangli2025TGRRG} achieves the highest BLEU-1 score, while MultiP-R2Gen~\cite{Chen2025MultiPR2Gen} performs best on BLEU-2, BLEU-3, and BLEU-4. DTrace~\cite{Ye2026DTrace} obtains the highest ROUGE-L score, whereas METransformer~\cite{Wang2023METransformer} and BootstrappingLLM~\cite{liu2024bootstrapping} achieve the best CIDEr and METEOR scores, respectively. Compared with the conference version, MambaXray-PRB-B slightly improves ROUGE-L, METEOR, and CIDEr over MambaXray-VL-Base~\cite{wang2025cxpmrg}, with CIDEr increasing from 0.206 to 0.231. MambaXray-PRB-L also improves METEOR from 0.167 to 0.169, while its remaining NLG scores are slightly lower than those of MambaXray-VL-Large~\cite{wang2025cxpmrg}. In terms of clinical efficacy, as reported in Table~\ref{tab:ce_comparison}, MambaXray-PRB-L achieves Precision, Recall, and F1 scores of 0.375, 0.334, and 0.348, respectively, with its Recall and F1 ranking second among the compared methods. The GREEN~\cite{ostmeier2024green} evaluation in Table~\ref{tab:green_metrics} provides a complementary assessment of clinical consistency. MambaXray-PRB-B achieves the lowest total number of clinically significant errors, reducing the aggregate error count to 3.564 compared with 4.533 for AM-MRG~\cite{wang2026ammrg} and 4.541 for R2GenCSR~\cite{wang2024r2gencsr}, while also obtaining the highest number of matched findings at 2.147. In particular, it substantially reduces errors in categories (a) and (b) to 0.057 and 0.097, respectively. However, limitations persist regarding clinical errors in categories (c) through (f). In summary, MambaXray-PRB remains competitive in terms of traditional report generation metrics and demonstrates strong performance across several clinically oriented metrics. These observations align with the rationale behind the revised post-training phase, which explicitly links disease-relevant image regions to the reasoning sequence, thereby influencing report generation.

Interestingly, our Base model outperforms its Large counterpart on several conventional NLG metrics, which may be related to their different reasoning budgets. The compact reasoning sequence of the Base model encourages attention to salient clinical findings and visual evidence. In contrast, the Large model employs a longer chain of thought to accommodate its higher-capacity visual representations and support more detailed multimodal reasoning. A longer reasoning sequence can capture more findings and incorporate richer localized visual evidence; however, it may also introduce redundant or weakly relevant information, potentially compromising the conciseness and standardization of radiology reports. This may partially explain the better lexical matching performance of MambaXray-PRB-B on MIMIC-CXR.

\noindent $\bullet$ \textbf{Results on IU X-ray Dataset.~}
On IU X-ray~\cite{demner2016iuxray}, MambaXray-PRB maintains competitive performance against existing report-generation methods and provides modest improvements over the conference version. Among the compared approaches, TGRRG~\cite{Guangli2025TGRRG} achieves the highest BLEU-1, BLEU-2, and ROUGE-L scores, while DTrace~\cite{Ye2026DTrace} and R2GenCSR~\cite{wang2024r2gencsr} obtain the best BLEU-3 and BLEU-4 scores, respectively. PromptMRG~\cite{Jin2024PromptMRG} achieves the highest METEOR score, whereas DCL~\cite{Li2023DCL} performs best on CIDEr. Compared with MambaXray-VL-Base~\cite{wang2025cxpmrg}, MambaXray-PRB-B improves BLEU-1/2/3/4 from 0.479/0.322/0.236/0.179 to 0.492/0.332/0.243/0.187. MambaXray-PRB-L similarly improves BLEU-2/3/4 from 0.330/0.241/0.185 to 0.336/0.250/0.195, together with small gains in ROUGE-L and CIDEr from 0.371/0.524 to 0.377/0.527. These results show that the revised framework retains competitive report-generation capability and yields modest improvements over MambaXray-VL~\cite{wang2025cxpmrg}, particularly in higher-order BLEU metrics. More importantly, the new post-training stage explicitly couples the reasoning sequence with localized disease-relevant patches before generating the report, enabling the model to associate clinical statements with corresponding visual evidence rather than relying solely on globally encoded image representations.

\noindent $\bullet$ \textbf{Results on CheXpert Plus Dataset.~}
Table~\ref{tab:XraybaseAlgorithms} reports the comparison on CheXpert Plus~\cite{chambon2024chexpertPLUS}. MambaXray-PRB achieves competitive performance across both NLG and clinical efficacy metrics. For NLG evaluation, MambaXray-VL-Large retains the highest BLEU-4 score of 0.112, while AM-MRG~\cite{wang2026ammrg} achieves the highest ROUGE-L and CIDEr scores of 0.282 and 0.221, and MultiP-R2Gen~\cite{Chen2025MultiPR2Gen} obtains the highest METEOR score of 0.269. MambaXray-PRB-L achieves a BLEU-4 score of 0.109, ranking second among the compared methods, while attaining competitive ROUGE-L, METEOR, and CIDEr scores of 0.274, 0.155, and 0.134. Compared with MambaXray-VL-Base, MambaXray-PRB-B improves BLEU-4, ROUGE-L, METEOR, and CIDEr from 0.105/0.267/0.149/0.117 to 0.108/0.271/0.153/0.130, respectively. The Large variant exhibits slightly lower NLG scores than MambaXray-VL-Large, with only small differences across the four metrics.

More pronounced improvements are observed in clinical efficacy. MambaXray-PRB-L achieves the best F1 score of 0.338 and the second-highest Precision of 0.387. Its Recall of 0.317 ranks third, while remaining highly competitive, trailing the best-performing MambaXray-VL-Large~\cite{wang2025cxpmrg} by only 0.002. MambaXray-PRB-B also consistently improves over MambaXray-VL-Base~\cite{wang2025cxpmrg}, increasing Precision, Recall, and F1 from 0.333/0.264/0.273 to 0.359/0.285/0.303. The GREEN~\cite{ostmeier2024green} results in Table~\ref{tab:green_metrics} further show that MambaXray-PRB-B achieves the lowest total number of clinically significant errors and the highest number of matched findings, with values of 3.428 and 1.805, respectively. It also obtains the lowest error counts in categories (a) and (b), at 0.083 and 0.048. Its GREEN score reaches 0.272, ranking second only to AM-MRG~\cite{wang2026ammrg} at 0.281, although its error counts in categories (c)--(f) remain higher than those of several competing methods. Overall, these results show that the proposed post-training strategy introduces evidence-grounded reasoning while retaining competitive linguistic quality and improving several clinically relevant evaluation measures. By explicitly incorporating localized disease-related patches into the reasoning sequence, MambaXray-PRB establishes a more direct connection between visual evidence, diagnostic reasoning, and the generated report than the conference version.

Table~\ref{tab:xrayLLMs} compares 25 representative LLMs and VLMs under a unified R2Gen-GPT-based evaluation setting. The results reveal clear differences among the evaluated models across lexical and clinical metrics. DeepSeek-R1-Distill-Llama-8B achieves the highest BLEU-4 score of 0.107, while Orca-2-7B~\cite{mitra2023orca} and Llama-2-13B~\cite{touvron2023llama2} obtain the best METEOR and CIDEr scores of 0.161 and 0.214, respectively. Vicuna-V1.5~\cite{zheng2023vicuna} remains highly competitive, ranking second on BLEU-4, ROUGE-L, CIDEr, and METEOR, with its METEOR score trailing the best-performing model by only 0.001. Notably, MedGemma-1.5-4B-IT~\cite{sellergren2026medgemma15} achieves the best ROUGE-L score of 0.278 and substantially outperforms the other evaluated models on the clinical efficacy metrics, reaching precision, recall, and F1 scores of 0.387, 0.299, and 0.309, respectively. In contrast, several general-purpose VLMs, including InternVL-2~\cite{chen2023internvl}, MiniCPM-V2.5~\cite{yao2024minicpmv}, and Qwen3-VL~\cite{yang2025qwen3}, exhibit relatively limited performance on this task. These comparisons suggest that model scale alone does not determine MRG performance; the pre-training characteristics and domain compatibility of the language or vision-language model play an important role in generating clinically relevant radiology reports.

\subsection{Parameter Analysis}
 
\begin{table} 
  \centering
  \caption{Performance under different maximum CoT lengths on the CheXpert Plus dataset using the base model. The maximum number of patch indices associated with each disease is fixed at 12.}
  \label{tab:cot_length}
  \scriptsize
  \setlength{\tabcolsep}{3pt}
  \renewcommand{\arraystretch}{1.05}
  \resizebox{\columnwidth}{!}{
  \begin{tabular}{c|ccccccc}
    \hline
    Maximum CoT Length & B4 & R-L & M & C & P & R & F1 \\
    \hline
    300 & 0.096 & 0.262 & 0.146 & 0.113 & 0.341 & 0.286 & 0.300 \\
    500 & 0.101 & 0.264 & 0.148 & 0.120 & 0.334 & 0.280 & 0.293 \\
    700 & 0.096 & 0.262 & 0.146 & 0.118 & 0.34 & 0.283 & 0.297 \\
    \hline
  \end{tabular}
  }
\end{table}

\begin{table} 
  \centering
  \caption{Performance of the base model on the CheXpert Plus dataset under different maximum numbers of patch indices associated with each disease. The maximum CoT length is fixed to 300 tokens.}
  \label{tab:patch_number}
  \scriptsize
  \setlength{\tabcolsep}{3pt}
  \renewcommand{\arraystretch}{1.05}
  \resizebox{\columnwidth}{!}{
  \begin{tabular}{c|ccccccc}
    \hline
    Maximum Number of Patches & B4 & R-L & M & C & P & R & F1 \\
    \hline
    6  & 0.096 & 0.262 & 0.146 & 0.114 & 0.339 & 0.281 & 0.295 \\
    12 & 0.096 & 0.262 & 0.146 & 0.113 & 0.341 & 0.286 & 0.300 \\
    18 & 0.097 & 0.261 & 0.147 & 0.113 & 0.329 & 0.275 & 0.290 \\
    \hline
  \end{tabular}
  }
\end{table}

\newcommand{\yes}{\textcolor{SeaGreen4}{\ding{51}}}
\newcommand{\no}{\textcolor{DarkRed}{\ding{55}}}

\begin{table*} 
\caption{Ablation study of the major components of our framework using the base model on the CheXpert Plus dataset.
\textbf{Vim-IN1K} denotes initialization with ImageNet-1K pre-trained weights;
\textbf{Vim-PTD} denotes initialization with weights pre-trained on 1.27 million X-ray images;
\textbf{MAE} and \textbf{ARG} denote masked auto-encoding and auto-regressive generation for visual pre-training, respectively;
\textbf{CTL} denotes X-ray--report contrastive learning;
\textbf{SFT} denotes supervised fine-tuning for report generation;
\textbf{Text-CoT} denotes LLM-generated textual chain-of-thought supervision,
and \textbf{MM-CoT} denotes the proposed multimodal chain-of-thought augmented with disease-related image patch indices.}
\label{tab:ablation_study}
\resizebox{\linewidth}{!}{
\begin{tabular}{l|cccccccc|ccccccc}
\hline \toprule [0.5 pt] 
\multirow{2}{*}{\textbf{Index}} & \multirow{2}{*}{\textbf{Vim-IN1K}} & \multirow{2}{*}{\textbf{Vim-PTD}} & \multirow{2}{*}{\textbf{MAE}} & \multirow{2}{*}{\textbf{ARG}} & \multirow{2}{*}{\textbf{CTL}} & \multirow{2}{*}{\textbf{SFT}} & \multirow{2}{*}{\textbf{Text-CoT}} & \multirow{2}{*}{\textbf{MM-CoT}} & \multicolumn{7}{c}{CheXpert Plus}  \\ \cline{10-16} 
 & & & & & & & & & \textbf{B4} & \textbf{R-L} &  \textbf{M}  & \textbf{C} & \textbf{P} & \textbf{R} & \textbf{F1} \\ \hline
$\#01$ & \no & \no & \no & \no & \no & \yes & \no & \no & 0.101 & 0.266 & 0.145 & 0.123 & 0.315 & 0.244 & 0.260 \\
\hline 
$\#02$ & \yes & \yes & \yes & \no &\no & \yes & \no & \no & 0.094 & 0.257 & 0.140 & 0.104 & 0.249 & 0.194 &0.208 \\
$\#03$ & \yes & \yes & \no & \yes & \no & \yes & \no & \no & 0.089 & 0.247 & 0.134 & 0.089 & 0.279 & 0.240 & 0.248 \\
\hline 
$\#04$ & \yes & \no & \no & \yes & \no & \yes & \no & \no & 0.090 & 0.249 & 0.132 & 0.103 & 0.262 & 0.195 & 0.209 \\
$\#05$ & \yes & \yes & \no & \yes & \no & \yes & \no & \no & 0.093 & 0.254 & 0.138 & 0.102 & 0.297 & 0.226 & 0.241 \\
$\#06$ & \yes & \yes & \no & \yes & \yes & \yes & \no & \no & 0.105 & 0.267 & 0.149 & 0.117 & 0.333 & 0.264 & 0.273 \\
\hline 
$\#07$ & \yes & \yes & \no & \yes & \yes & \yes & \yes & \no & 0.107 & 0.270 & 0.152 & 0.128 & 0.342 & 0.264 & 0.278 \\
$\#08$ & \yes & \yes & \no & \yes & \yes & \yes & \yes & \yes & \textbf{0.108} & \textbf{0.271} & \textbf{0.153} & \textbf{0.130} & \textbf{0.359} & \textbf{0.285} & \textbf{0.303} \\
\hline \toprule [0.5 pt] 
\end{tabular}
}
\end{table*} 

We further investigate two key configurations associated with the proposed MM-CoT, including the maximum CoT sequence length and the maximum number of patch indices incorporated into the reasoning sequence for each disease. All experiments are conducted using the base model on the CheXpert Plus dataset.

\noindent $\bullet$ \textbf{Effect of Maximum CoT Length.}
We first study the influence of the token budget allocated to CoT generation. As shown in Table~\ref{tab:cot_length}, we vary the maximum CoT length among 300, 500, and 700 tokens while fixing the maximum number of patch indices to 12. Increasing the maximum length from 300 to 500 tokens slightly improves the conventional language-generation metrics, with BLEU-4, ROUGE-L, METEOR, and CIDEr increasing from 0.096/0.262/0.146/0.113 to 0.101/0.264/0.148/0.120, respectively. However, this does not lead to better clinical performance. In contrast, the setting of 300 tokens achieves the highest precision, recall, and F1 scores of 0.341, 0.286, and 0.300, respectively. Further extending the maximum length to 700 tokens provides no consistent benefit.

These results suggest that a longer reasoning sequence does not necessarily provide more effective supervision for report generation. Although a larger token budget allows more detailed reasoning content to be generated, excessively long CoTs may introduce redundant descriptions that contribute little to learning disease-relevant visual representations. Considering both clinical performance and generation efficiency, we therefore set the maximum input CoT length to 300 tokens in the remaining CheXpert Plus experiments.

\noindent $\bullet$ \textbf{Effect of Maximum Patch Number.}
We next analyze the effect of varying the maximum number of patch indices associated with each disease in MM-CoT. As reported in Table~\ref{tab:patch_number}, the maximum CoT length is fixed at 300 tokens. Across three separate experiments, the maximum number of patch indices per disease label is uniformly set to 6, 12, and 18, respectively. The conventional text-generation metrics remain relatively stable under different settings. In contrast, the clinical metrics exhibit a clearer dependence on the number of selected patches. Increasing the maximum patch number from 6 to 12 improves precision, recall, and F1 from 0.339/0.281/0.295 to 0.341/0.286/0.300, respectively. When the number is further increased to 18, the clinical performance decreases to 0.329/0.275/0.290, despite marginal changes in BLEU-4 and METEOR.

The results indicate that incorporating an appropriate amount of regional information is important for constructing effective MM-CoT supervision. A small patch budget may omit image regions associated with relevant abnormalities, whereas including too many patches can introduce less discriminative or redundant regional information and weaken the correspondence between diagnostic reasoning and salient radiographic evidence. Among the evaluated configurations, a maximum of 12 patches provides the best balance and is therefore adopted as the default setting.

\subsection{Ablation Study}

\noindent $\bullet$ \textbf{Auto-regressive Generation vs. Masked Reconstruction for Visual Pre-training.}
We first investigate the influence of different visual pre-training objectives by comparing masked auto-encoding (MAE) and auto-regressive generation (ARG). As reported in rows $\#02$ and $\#03$ of Table~\ref{tab:ablation_study}, replacing MAE with ARG leads to higher clinical precision, recall, and F1 scores, increasing them from 0.249/0.194/0.208 to 0.279/0.240/0.248, respectively. In particular, the recall and F1 scores are improved by approximately 23.7\% and 19.2\%. Although the conventional language-generation metrics are slightly lower with ARG in this comparison, its clear advantage on the clinical metrics indicates that auto-regressive visual pre-training is more effective at encoding disease-related information that is important for radiology report generation. Therefore, ARG is adopted as the visual pre-training objective in the subsequent experiments.

\noindent $\bullet$ \textbf{Effect of X-ray-specific Visual Pre-training.}
We further examine whether pre-training the vision encoder on large-scale X-ray images provides additional benefits beyond ImageNet-1K initialization. Rows $\#04$ and $\#05$ differ only in whether the X-ray pre-trained Vim weights are used. Introducing Vim-PTD improves BLEU-4, ROUGE-L, and METEOR from 0.090/0.249/0.132 to 0.093/0.254/0.138, while the CIDEr score remains nearly unchanged. More importantly, the P/R/F1 scores increase from 0.262/0.195/0.209 to 0.297/0.226/0.241. The gains in recall and F1 reach approximately 15.9\% and 15.3\%, respectively. These results suggest that domain-specific pre-training enables the visual encoder to learn representations that are better aligned with radiographic abnormalities than those obtained solely from natural-image pre-training.

\noindent $\bullet$ \textbf{Contribution of X-ray--Report Contrastive Learning.}
To evaluate the role of cross-modal alignment, we compare rows $\#05$ and $\#06$ by introducing the contrastive learning objective between X-ray images and their corresponding reports. CTL consistently improves all evaluated metrics. Specifically, BLEU-4, ROUGE-L, METEOR, and CIDEr increase from 0.093/0.254/0.138/0.102 to 0.105/0.267/0.149/0.117, respectively. Similar improvements are observed for the clinical metrics, where the F1 score rises from 0.241 to 0.273 and the recall increases from 0.226 to 0.264. The consistent gains demonstrate that explicitly aligning visual and textual representations facilitates the transfer of clinically relevant information from images to report generation.

\definecolor{ReportYellow}{RGB}{255,255,140}
\definecolor{CoTCyan}{RGB}{160,235,245}
\definecolor{PatchPink}{RGB}{255,185,210}

\newcommand{\MatchingReport}[1]{%
    \sethlcolor{ReportYellow}\hl{#1}%
}

\newcommand{\MatchingCoT}[1]{%
    \sethlcolor{CoTCyan}\hl{#1}%
}

\newcommand{\VisualPatch}[1]{%
    \sethlcolor{PatchPink}\hl{#1}%
}

\newcommand{\CoTLess}{%
    \sethlcolor{CoTCyan}\hl{\mbox{\textless}}%
}

\newcommand{\CoTGreater}{%
    \sethlcolor{CoTCyan}\hl{\mbox{\textgreater}}%
}

\newcolumntype{C}[1]{%
    >{\centering\arraybackslash}m{#1}%
}

\newcolumntype{J}[1]{%
    >{\justifying\arraybackslash
      \hyphenpenalty=50
      \exhyphenpenalty=50
      \emergencystretch=2em}m{#1}%
}

\setlength{\fboxsep}{0.3pt}

\newcommand{\PatchImg}[1]{%
    \colorbox{PatchPink}{%
        \raisebox{-0.15\height}{%
            \includegraphics[height=2.0ex]{#1}%
        }%
    }%
}

\begin{figure*}[t]
\centering

\resizebox{\textwidth}{!}{%
\begin{tabular}{
    C{4cm}
    J{7cm}
    J{7cm}
    J{10cm}
}

\toprule

\multicolumn{1}{c}{\textbf{Image}} &
\multicolumn{1}{c}{\textbf{Ground Truth}} &
\multicolumn{1}{c}{\textbf{Generated Report}} &
\multicolumn{1}{c}{\textbf{Multimodal CoT}}
\\

\midrule


\includegraphics[width=3cm]{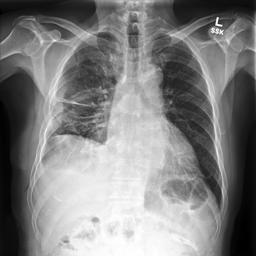}

&
\MatchingReport{the cardiomediastinal silhouette is normal .}
\MatchingReport{small right pleural effusion with loculated component decreased compared to chest x-ray 2021 january 19th .}
the left lung is clear .
\MatchingReport{no acute bony abnormalities .}
\MatchingReport{there is persistent elevation of the right hemidiaphragm .}
skin staples over midline abdomen .

&
\MatchingReport{the cardiomediastinal silhouette is unchanged .}
\MatchingReport{there is redemonstration of a small right pleural effusion and elevation of the right hemidiaphragm .}
there is improved aeration of the left lung base compared to the prior study .
there is no evidence of pulmonary edema or pneumothorax .
\MatchingReport{there are no acute osseous abnormalities .}
surgical clips project over the right upper quadrant of the abdomen .

&
Alright, let's break this down step by step.
First, I notice the cardiomediastinal silhouette appears stable---no signs of
\CoTLess\MatchingCoT{Enlarged Cardiomediastinum}\CoTGreater
or
\CoTLess\MatchingCoT{Cardiomegaly}\CoTGreater,
which is reassuring.
Moving to the right hemithorax, there's a persistent small
\CoTLess
\MatchingCoT{Pleural Effusion }
\PatchImg{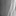}
\PatchImg{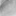}
\PatchImg{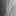}
\PatchImg{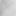}
\PatchImg{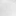}
\PatchImg{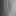}
\PatchImg{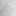}
\PatchImg{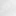}
\PatchImg{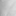}
\PatchImg{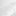}
\PatchImg{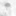}
\CoTGreater
with associated elevation of the right hemidiaphragm.
This seems unchanged from prior imaging, so no acute progression.
Now, focusing on the right lung base, there's a linear
\CoTLess
\MatchingCoT{Lung Opacity }
\PatchImg{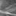}
\PatchImg{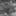}
\PatchImg{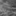}
\PatchImg{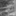}
\PatchImg{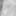}
\PatchImg{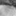}
\PatchImg{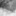}
\PatchImg{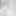}
\PatchImg{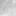}
\CoTGreater

\\

\midrule


\includegraphics[width=3cm]{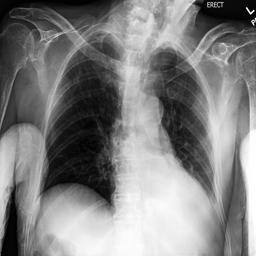}

&
an ap upright radiograph of the chest was obtained .
\MatchingReport{there is a stable left lower lobe consolidation which may represent atelectais or infection .}
\MatchingReport{a small left pleural effusion is unchanged .}
the right lung is grossly unremarkable .
\MatchingReport{the cardiomediastinal silhouette is stable .}
\MatchingReport{the pulmonary vasculature is within normal limits .}
there is moderate osteopenia of the visualized osseous structures .

&
single frontal view of the chest demonstrates interval removal of the right internal jugular central venous catheter .
\MatchingReport{there is redemonstration of a left retrocardiac opacity and blunting of the left costophrenic angle which may represent atelectasis or consolidation .}
there is no evidence of pneumothorax .
\MatchingReport{the cardiomediastinal silhouette is unremarkable .}
\MatchingReport{the pulmonary vasculature is within normal limits .}
the osseous structures are unremarkable .

&
Alright, let's break this down step by step.
First, I notice the overall
\CoTLess
\MatchingCoT{Low Lung Volumes}
\CoTGreater,
which could suggest suboptimal inspiration or underlying pathology.
Moving to the left hemithorax, there's a subtle
\CoTLess
\MatchingCoT{Pleural Effusion }
\PatchImg{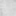}
\PatchImg{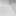}
\PatchImg{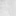}
\PatchImg{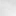}
\PatchImg{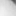}
\PatchImg{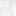}
\PatchImg{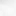}
\PatchImg{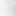}
\PatchImg{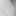}
\PatchImg{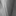}
\PatchImg{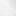}
\PatchImg{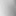}
\CoTGreater
---it's small but worth noting.
Now, focusing on the retrocardiac region, there's a persistent
\CoTLess
\MatchingCoT{Lung Opacity }
\PatchImg{patches291.png}
\PatchImg{patches292.png}
\PatchImg{patches2103.png}
\PatchImg{patches2104.png}
\PatchImg{patches2105.png}
\PatchImg{patches2115.png}
\PatchImg{patches2116.png}
\PatchImg{patches2117.png}
\PatchImg{patches2118.png}
\PatchImg{patches2119.png}
\PatchImg{patches2129.png}
\PatchImg{patches2130.png}
\CoTGreater
that stands out.

\\

\midrule


\includegraphics[width=3cm]{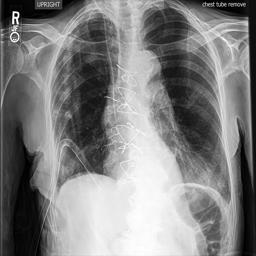}

&
removal more superior of 2 right-sided chest tubes .
\MatchingReport{right ij and second chest tube remain in place .}
\MatchingReport{small right apical pneumothorax is identified .}
\MatchingReport{increasing parenchymal opacity in left base with persistent small left pleural effusion .}

&
\MatchingReport{interval placement of a right internal jugular central venous catheter with tip in the lower svc .}
\MatchingReport{unchanged positioning of median sternotomy wires mediastinal drain}
and right chest tube .
\MatchingReport{small right apical pneumothorax again seen .}
\MatchingReport{persistent left basilar opacity}
likely atelectasis or consolidation .
cardiomediastinal silhouette is unchanged .
no acute bony abnormalities .

&
First, I notice the
\CoTLess
\MatchingCoT{Support Devices }
\PatchImg{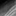}
\PatchImg{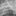}
\PatchImg{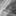}
\PatchImg{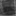}
\PatchImg{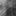}
\PatchImg{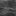}
\PatchImg{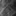}
\PatchImg{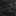}
\PatchImg{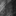}
\PatchImg{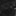}
\PatchImg{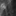}
\PatchImg{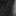}
\CoTGreater
---specifically, the right internal jugular central venous catheter and
mediastinal drains appear unchanged from prior imaging.
Moving to the lungs, there's a small
\CoTLess
\MatchingCoT{Pneumothorax }
\PatchImg{patches328.png}
\PatchImg{patches329.png}
\PatchImg{patches340.png}
\PatchImg{patches341.png}
\PatchImg{patches352.png}
\PatchImg{patches353.png}
\PatchImg{patches364.png}
\PatchImg{patches365.png}
\CoTGreater
in the right apex, which seems stable compared to prior studies.
Hmm, the bibasilar opacities are still present, likely representing
\CoTLess
\MatchingCoT{Atelectasis}
\CoTGreater
or
\CoTLess
\MatchingCoT{Consolidation}
\CoTGreater,
given their location and persistence over time.
Additionally, there's a subtle
\CoTLess
\MatchingCoT{Pleural Effusion }
\PatchImg{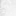}
\PatchImg{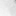}
\PatchImg{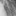}
\CoTGreater

\\

\midrule


\includegraphics[width=3cm]{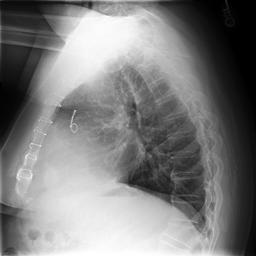}

&
\MatchingReport{pa and lateral upright views of the chest demonstrate clear lung fields bilaterally .}
\MatchingReport{the heart is mildly enlarged .}
the thoracic aorta demonstrates mild prominence of the ascending aorta .
\MatchingReport{there are no pleural effusions .}
\MatchingReport{post surgical changes of prior coronary revascularization are seen .}
anterior wedging is seen of the vertebral body at the thoracolumbar junction .

&
\MatchingReport{sternotomy wires and mediastinal clips are redemonstrated .}
\MatchingReport{the cardiomediastinal silhouette is mildly enlarged .}
\MatchingReport{there is no evidence of focal consolidation pulmonary edema or pleural effusion .}
degenerative changes of the thoracic spine are again noted as well as mild scoliosis of the thoracic spine .

&
Alright, let's break this down step by step.
First, I notice the
\CoTLess
\MatchingCoT{Support Devices}
\CoTGreater
---specifically, the sternotomy wires and mediastinal clips---which are
clearly visible and unchanged from prior studies.
Moving to the cardiomediastinum, there's mild
\CoTLess
\MatchingCoT{Cardiomegaly }
\PatchImg{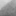}
\PatchImg{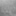}
\PatchImg{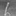}
\PatchImg{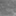}
\PatchImg{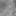}
\PatchImg{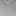}
\PatchImg{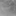}
\PatchImg{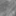}
\PatchImg{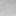}
\PatchImg{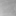}
\PatchImg{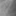}
\PatchImg{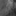}
\CoTGreater,
but it's not severe enough to warrant concern.
Now, focusing on the lungs: they appear clear bilaterally, with no signs of
\CoTLess
\MatchingCoT{Lung Opacity}
\CoTGreater,
\CoTLess
\MatchingCoT{Edema}
\CoTGreater,
or other abnormalities.
The costophrenic angles are sharp, which rules out
\CoTLess
\MatchingCoT{Pleural Effusion}
\CoTGreater.
Overall, the findings are stable compared to prior studies.

\\

\bottomrule

\end{tabular}%
}

\caption{Qualitative examples of generated radiology reports and multimodal chain-of-thought (MM-CoT) reasoning. Yellow highlights denote corresponding clinical findings in the ground-truth and generated reports. Cyan highlights indicate clinical concepts identified during MM-CoT reasoning, while pink-framed image patches denote the corresponding visual evidence.}
\label{fig:visual_report}
\end{figure*}

\definecolor{ReportYellow}{RGB}{255,255,140}
\definecolor{CoTCyan}{RGB}{160,235,245}
\definecolor{PatchPink}{RGB}{255,185,210}




%

\newcolumntype{C}[1]{%
    >{\centering\arraybackslash}m{#1}%
}

\newcolumntype{J}[1]{%
    >{\justifying\arraybackslash
      \hyphenpenalty=50
      \exhyphenpenalty=50
      \emergencystretch=2em}m{#1}%
}

\setlength{\fboxsep}{0.3pt}

\begin{figure*}[t]
\centering

\resizebox{\textwidth}{!}{%
\begin{tabular}{
    C{5cm}
    J{8cm}
    J{8cm}
    J{8cm}
}

\toprule

\multicolumn{1}{c}{\textbf{Image}} &
\multicolumn{1}{c}{\textbf{Ground Truth}} &
\multicolumn{1}{c}{\textbf{Ours}} &
\multicolumn{1}{c}{\textbf{R2GenGPT}}
\\

\midrule


\includegraphics[width=3cm]{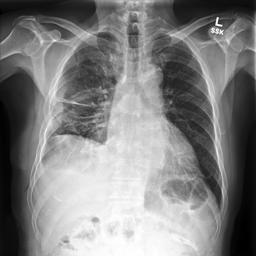}

& 
\VisualPatch{unchanged right internal jugular central venous catheter}
\MatchingReport{prosthetic aortic valve}
epicardial pacer leads and
\VisualPatch{median sternotomy wires} .
\MatchingReport{unchanged tiny right apical pneumothorax} .
persistent left lower lobe
\VisualPatch{atelectasis versus consolidation} .
\VisualPatch{unchanged small bilateral pleural effusions} .
unchanged mild cardiomegaly .

&
\MatchingReport{stable right ij central venous catheter}
\MatchingReport{prosthetic aortic valve}
and
\MatchingReport{sternotomy wires} .
interval removal of mediastinal drains .
\MatchingReport{persistent small right apical pneumothorax} .
persistent mild pulmonary edema .
persistent bibasilar
\MatchingReport{opacities which may reflect atelectasis or consolidation} .
\MatchingReport{small bilateral pleural effusions} .
unchanged cardiomediastinal silhouette .
no acute osseous abnormalities .

&
\MatchingCoT{unchanged positioning of right internal jugular central venous catheter}
mediastinal drainage tube
\MatchingCoT{sternotomy wires}
and multiple thoracic drains .
interval removal of swan-ganz catheter .
stable cardiomediastinal silhouette .
\MatchingCoT{persistent small bilateral pleural effusions} .
persistent bibasilar
\MatchingCoT{opacities likely atelectasis versus consolidation} .
mild pulmonary edema .
no pneumothorax .

\\

\midrule


\includegraphics[width=3cm]{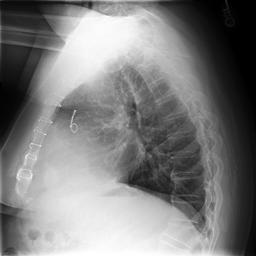}

&
\VisualPatch{interval placement of a right internal jugular central venous catheter}
terminating 43 cm below the carina in the lower superior vena cava .
persistent
\MatchingReport{bibasilar}
\VisualPatch{opacities likely atelectasis versus consolidation} .
\MatchingReport{small bilateral pleural effusions} .
\VisualPatch{no pneumothorax} .
\MatchingReport{redemonstration of marked enlargement of the thoracic aorta with associated extensive widening of the mediastinum} .

&
\MatchingReport{interval placement of a right internal jugular central venous catheter}
with the tip in the mid svc .
stable positioning of mediastinal drains and prosthetic aortic valve .
stable cardiomediastinal silhouette with
\MatchingReport{widening of the superior mediastinum consistent with known aortic aneurysm} .
mild pulmonary edema .
\MatchingReport{small bilateral pleural effusions} .
\MatchingReport{bibasilar opacities likely reflect atelectasis versus consolidation} .
\MatchingReport{no pneumothorax} .

&
ap semierect view of the chest demonstrates
\MatchingCoT{interval placement of a right internal jugular central venous catheter}
with the tip in the mid svc .
\MatchingCoT{there is no pneumothorax} .
the cardiomediastinal silhouette is enlarged .
there is mild pulmonary edema .
there is retrocardiac
\MatchingCoT{opacification which may represe likely atelectasis or consolidation} .
mild pulmonary edema .

\\

\bottomrule

\end{tabular}%
}

\caption{X-ray images and their corresponding ground-truths, along with the output of our model and R2GenGPT model generation reports on the CheXpert Plus dataset. Matching sentences in our report are highlighted in yellow, R2GenGPT matching sentences are highlighted in cyan, and sentences matching by both models are highlighted in pink.}
\label{fig:visual_report_contrast}
\end{figure*}

\noindent $\bullet$ \textbf{Effect of Chain-of-Thought Generation.}
We next investigate whether introducing an additional reasoning generation task benefits X-ray report generation. It should be noted that the chain-of-thought (CoT) is introduced only in the third training stage and does not participate in either visual pre-training or X-ray--report contrastive learning. During post-training, report generation and CoT generation follow the same vision-to-language generation pipeline but are activated by different prompts. The language model is kept frozen, while the losses from both generation tasks are jointly used to optimize the visual encoder and the projection layer. 

As shown by rows $\#06$ and $\#07$ in Table~\ref{tab:ablation_study}, adding the textual CoT generation task improves most evaluation metrics. BLEU-4, METEOR, and CIDEr increase from 0.105/0.149/0.117 to 0.107/0.152/0.128, respectively, with a relative improvement of approximately 9.4\% on CIDEr. The clinical F1 score also increases from 0.273 to 0.278. These results indicate that jointly learning report generation and diagnostic reasoning provides additional supervision for adapting the visual representation to the frozen language model, thereby improving the downstream report generation performance.

\noindent $\bullet$ \textbf{Effect of Patch-aware Multimodal Chain-of-Thought.}
We further study whether incorporating disease-related spatial information into the CoT target can provide more effective supervision. For each training sample, the image patches associated with the abnormalities described in the report are identified during data preparation, and their patch indices are inserted into the corresponding reasoning sequence to construct the multimodal chain-of-thought (MM-CoT). Similar to Text-CoT, MM-CoT is generated only in the third stage through a dedicated prompt.

Rows $\#07$ and $\#08$ compare textual CoT with the proposed patch-aware MM-CoT. Introducing patch indices yields consistent improvements across all reported metrics. In particular, precision, recall, and F1 increase from 0.342/0.264/0.278 to 0.359/0.285/0.303, corresponding to relative gains of approximately 5.0\%, 8.0\%, and 9.0\%, respectively. Improvements are also observed in BLEU-4, ROUGE-L, METEOR, and CIDEr. Since the only difference between the two configurations is the inclusion of disease-related patch indices in the reasoning target, the results demonstrate that coupling textual reasoning with explicit abnormality localization provides more informative supervision for the shared visual representation. Consequently, the visual encoder and projection layer can better preserve clinically relevant regional information for subsequent language generation.

Overall, the complete configuration in row $\#08$ achieves the best performance across all reported metrics. These results demonstrate the complementary contributions of auto-regressive visual pre-training, X-ray--report contrastive alignment, and joint report--reasoning generation. In particular, enriching the reasoning target with disease-related patch indices introduces fine-grained spatial supervision without modifying the report generation pathway, leading to further improvements in clinical accuracy.

\subsection{Visualization}



Fig.~\ref{fig:visual_report} presents qualitative examples illustrating the correspondence among chest X-ray images, generated reports, and MM-CoT reasoning. The visualization shows that the model can associate clinically relevant findings with spatially meaningful visual evidence. In the first example, the generated report correctly describes a small right pleural effusion and an elevated right hemidiaphragm, while the corresponding image patches highlighted in the MM-CoT are concentrated in the lower-right thoracic region. In the second example, the left retrocardiac opacity and pleural abnormality are reflected in both the generated report and MM-CoT, with the associated visual evidence primarily located in the left lower lung and retrocardiac region. The third example further demonstrates the correspondence between a small right apical pneumothorax and image patches from the corresponding upper-right thoracic region. Similarly, in the fourth example, the visual evidence associated with cardiomegaly mainly covers the central cardiomediastinal region, consistent with the mildly enlarged cardiac silhouette described in the report. Overall, these examples indicate that the learned MM-CoT establishes meaningful associations between clinical concepts and spatially relevant visual evidence, providing additional interpretability for the visual representations learned by the framework. The reasoning sequence and radiology report are jointly modeled within the unified autoregressive generation process, enabling clinical reasoning to be explicitly grounded in localized image evidence. Nevertheless, these visualizations are qualitative and should be interpreted as evidence of cross-modal correspondence rather than precise lesion segmentation.

As illustrated in Fig.~\ref{fig:visual_report_contrast}, our method shows stronger agreement with the reference reports on several clinically salient findings in these qualitative examples, including pneumothorax, pleural effusion, and aortic/mediastinal abnormalities. Compared with R2GenGPT, our generated reports recover more reference-consistent clinical details, while both methods still exhibit occasional unsupported findings or imprecise anatomical localization.

\subsection{Limitation Analysis}  
Despite the promising results, several limitations remain. First, our multimodal Chain-of-Thought supervision is automatically constructed using an LLM and an auxiliary disease recognition model. Although this strategy enables scalable reasoning supervision, the generated reasoning traces and localized evidence may inherit errors from these models and should therefore be regarded as auxiliary supervision rather than definitive clinical reasoning. Second, the current visual grounding relies on a predefined set of thoracic abnormalities and discrete patch indices, which may be insufficient for subtle findings or abnormalities outside the predefined categories. Finally, our experiments are limited to publicly available chest X-ray datasets. Further evaluation across institutions and patient populations, together with radiologist assessment, is necessary to establish the robustness and clinical utility of the proposed framework.

\section{Conclusion and Future Works} \label{sec:Conclusion}
In this paper, we revisit X-ray radiology report generation from the perspectives of large-scale pre-training, cross-modal alignment, and evidence-grounded reasoning. We establish a comprehensive benchmark on CheXpert Plus to systematically evaluate representative radiology report generation methods and foundation models. Furthermore, we propose MambaXray-PRB, which progressively integrates auto-regressive X-ray visual pre-training, X-ray--report contrastive learning, and multimodal Chain-of-Thought post-training. By explicitly associating diagnostic reasoning with disease-relevant image patches, the proposed framework establishes a direct connection between visual evidence, intermediate reasoning, and generated reports. Extensive experiments on IU X-ray, MIMIC-CXR, and CheXpert Plus demonstrate competitive report generation performance and improvements in several clinically oriented metrics, while ablation studies validate the complementary contributions of the proposed components.

In future work, we plan to investigate expert-verified reasoning supervision and finer-grained visual grounding to further improve the reliability of multimodal reasoning. Cross-institutional validation and radiologist-centered evaluation will also be important for assessing robustness and clinical applicability. Moreover, extending the proposed paradigm to other medical imaging modalities and developing more efficient adaptive reasoning strategies represent promising directions for future research.

\small{ 
\bibliographystyle{IEEEtran}
\bibliography{reference}
}


\end{CJK}
\end{document}